\documentclass{article}

\usepackage{arxiv}

\usepackage[utf8]{inputenc}
\usepackage{color}
\usepackage{soul}
\usepackage{graphicx}
\usepackage{amsmath}
\usepackage{amssymb}
\usepackage{booktabs}
\usepackage{url}
\usepackage{xcolor}
\usepackage{xspace}
\graphicspath{ {figures/} }
\usepackage{booktabs} 
\usepackage[inline]{enumitem}
\usepackage{multirow}
\usepackage[normalem]{ulem}
\usepackage{float}
\usepackage{array}

\usepackage[pagebackref,breaklinks,colorlinks]{hyperref}
\usepackage{url}

\newcolumntype{C}[1]{>{\centering\arraybackslash}p{#1}}

\def\correspondingauthor{\thanks{Address for correspondence: \texttt{open-buildings-dataset@google.com}}}

\title{High-resolution building and road detection from Sentinel-2}
\author{
Wojciech Sirko, Emmanuel Asiedu Brempong, Juliana T. C. Marcos, Abigail Annkah \\ 
\bf{Abel Korme, Mohammed Alewi Hassen, Krishna Sapkota, Tomer Shekel,}\\
\bf{Abdoulaye Diack, Sella Nevo, Jason Hickey, John Quinn \correspondingauthor{}} \vspace{.3cm} \\
Google Research
}

\usepackage[bottom]{footmisc} 
\usepackage{graphicx}
\usepackage{hyperref}
\usepackage{subcaption}

\hypersetup{
  colorlinks,
  breaklinks=true,
  anchorcolor={blue!50!black},
  linkcolor={blue!50!black},
  citecolor={blue!50!black},
  urlcolor={blue!50!black}
}

\usepackage[capitalize]{cleveref}
\crefname{figure}{Fig.}{Figs.}
\Crefname{figure}{Figure}{Figures}
\crefname{section}{Sec.}{Secs.}
\Crefname{section}{Section}{Sections}
\Crefname{table}{Table}{Tables}
\crefname{table}{Tab.}{Tabs.}

\begin{document}
\maketitle

\begin{abstract}
Mapping buildings and roads automatically with remote sensing typically requires high-resolution imagery, which is expensive to obtain and often sparsely available. In this work we demonstrate how multiple 10 m resolution Sentinel-2 images can be used to generate 50 cm resolution building and road segmentation masks. This is done by training a `student' model with access to Sentinel-2 images to reproduce the predictions of a `teacher' model which has access to corresponding high-resolution imagery. While the predictions do not have all the fine detail of the teacher model, we find that we are able to retain much of the performance: for building segmentation we achieve 79.0\% mIoU, compared to the high-resolution teacher model accuracy of 85.5\% mIoU. We also describe two related methods that work on Sentinel-2 imagery: one for counting individual buildings which achieves $R^2 = 0.91$ against true counts and one for predicting building height with 1.5 meter mean absolute error. This work opens up new possibilities for using freely available Sentinel-2 imagery for a range of tasks that previously could only be done with high-resolution satellite imagery.
\end{abstract}

\section{Introduction}

Buildings and roads are important to map for a range of practical applications. Models that do this automatically using high-resolution (50 cm or better) satellite imagery are increasingly effective. However, this type of imagery is difficult to obtain: there is little control over revisit times, the cost can be prohibitive for larger analyses, and there may be limited or no historical imagery for a particular area. This rules out certain analyses, such as spatially comprehensive surveys of buildings and roads, or systematic study of changes over time, e.g. for studying urbanisation, economic or environmental changes.

\begin{figure}[b]
    \centering
    \includegraphics[width=\textwidth]{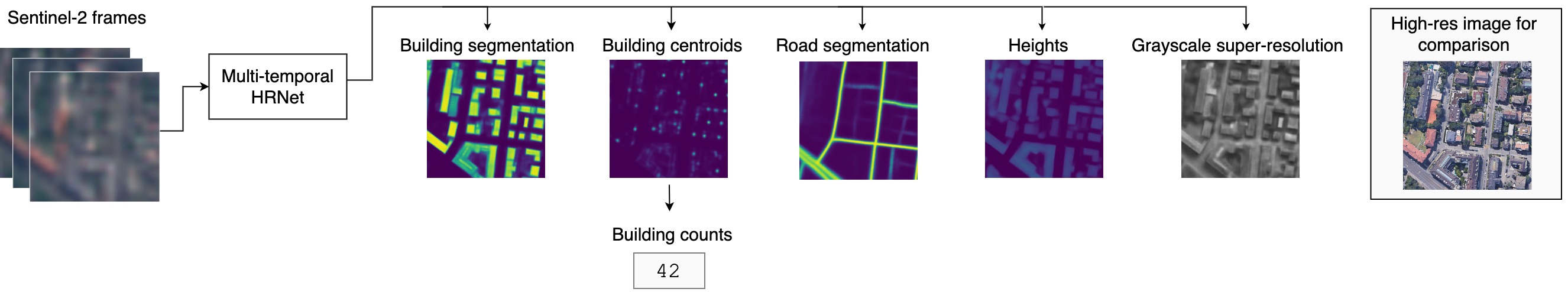}
    \caption{\label{fig:outputs}Example operation of our model, where multiple frames of low-resolution Sentinel-2 imagery are used to make a single frame of high-resolution predictions for a variety of output types. A high-resolution image of the same scene is shown for comparison.}
\end{figure}

The Sentinel-2 earth observation missions collect imagery globally every 2-5 days and depending on the band at up to 10 m ground resolution. This freely available data source is commonly used for the adjacent task of land cover mapping, and recent work on super-resolution with Sentinel-2 imagery has shown that a higher level of detail can be obtained than the native image resolution would suggest. This is intuitively possible because of the small variation in spatial position across successive Sentinel-2 image frames, caused by atmospheric disturbances, and even across bands in the same image due to sensor layout. A slightly different $10^2$ m$^2$ area is being captured each time, and hence a sequence of such images can be used to reconstruct finer detail than exists in any one frame. We review related work in Section \ref{sec:related}, which includes a number of experiments on obtaining 2.5m super-resolution from Sentinel-2.

\begin{figure}
  \centering
  \includegraphics[width=\textwidth]{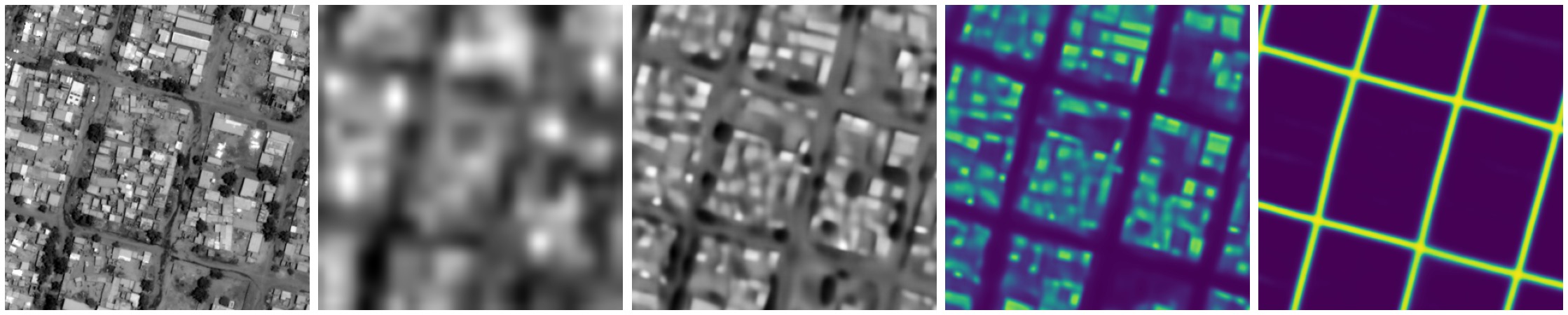}
  \includegraphics[width=\textwidth]{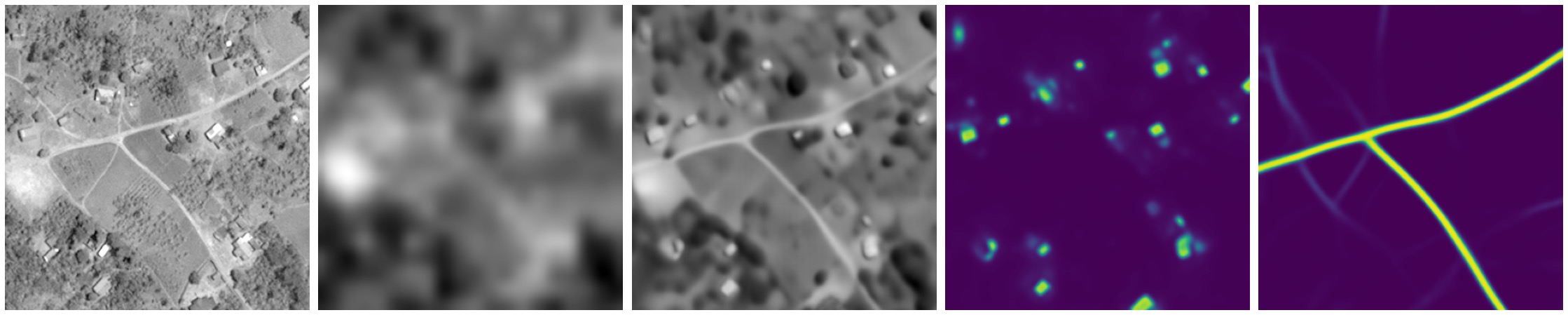}
    
  \begin{tabular}{*{5}{C{0.175\textwidth}}}
     \footnotesize 50 cm grayscale & \footnotesize Sentinel-2 grayscale & \footnotesize Sentinel-2 grayscale super-resolution & \footnotesize Building detection & \footnotesize Road detection \\
  \end{tabular}

  \caption{\label{fig:building-road-examples}Examples of building and road detection from Sentinel-2 imagery, each covering an area of $192^2$ m$^2$. The panels on the left show high-resolution satellite imagery of the scene for comparison; although Sentinel-2 imagery has much lower level of detail in each frame, we are able to predict fine-scale features of buildings and roads.}

\end{figure}

\begin{figure}[b]
    \centering
    \includegraphics[width=0.49\textwidth]{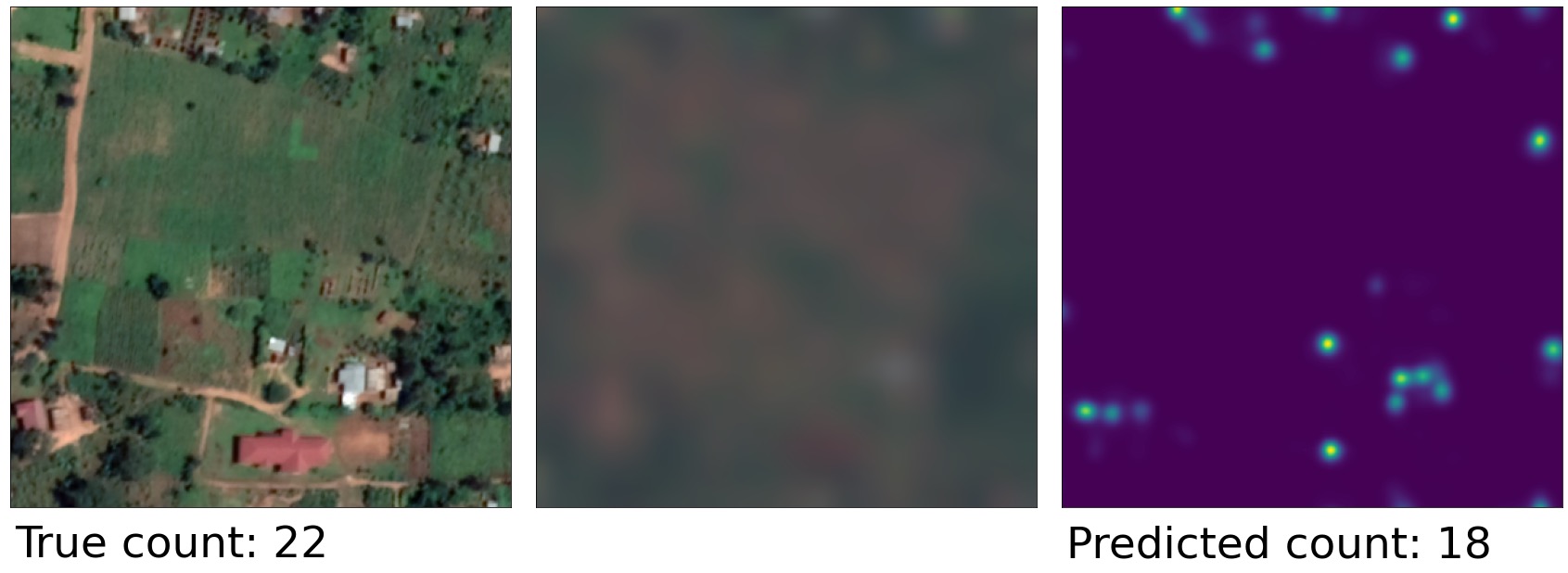}
    \hspace{.01\textwidth}
    \includegraphics[width=0.49\textwidth]{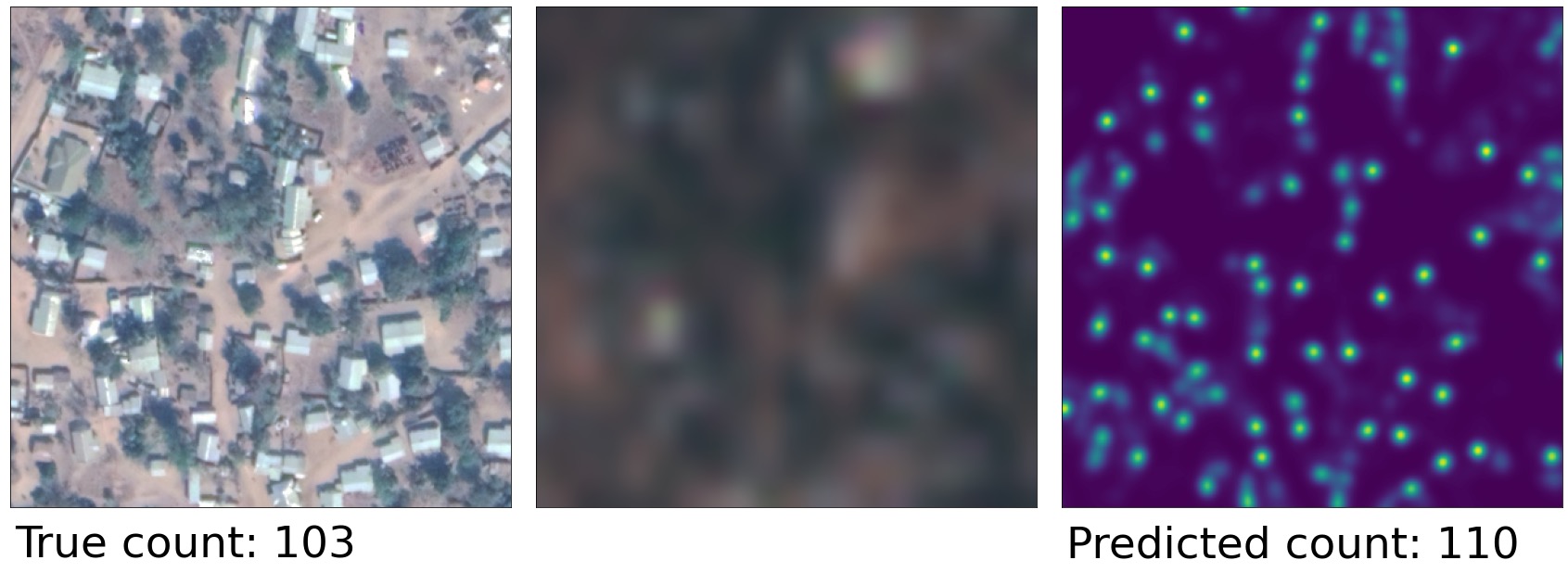}
    \caption{\label{fig:centroid-count-examples}Estimation of the number of buildings in a tile, based on predicting building centroids (left: high resolution image for comparison, centre: Sentinel-2 RGB; right: predicted centroid mask). This method can obtain $R^2 = 0.91$ with respect to true counts even though individual buildings cannot be discerned in the source imagery.}
\end{figure}

\begin{figure}[b]
    \centering
    \includegraphics[width=0.49\textwidth]{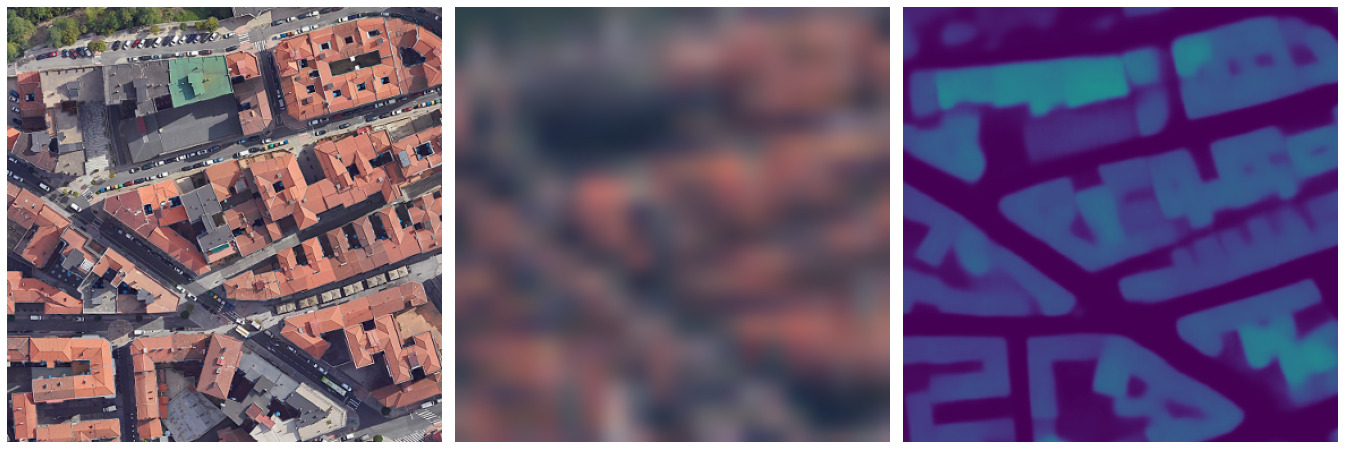}
    \hspace{.01\textwidth}
    \includegraphics[width=0.49\textwidth]{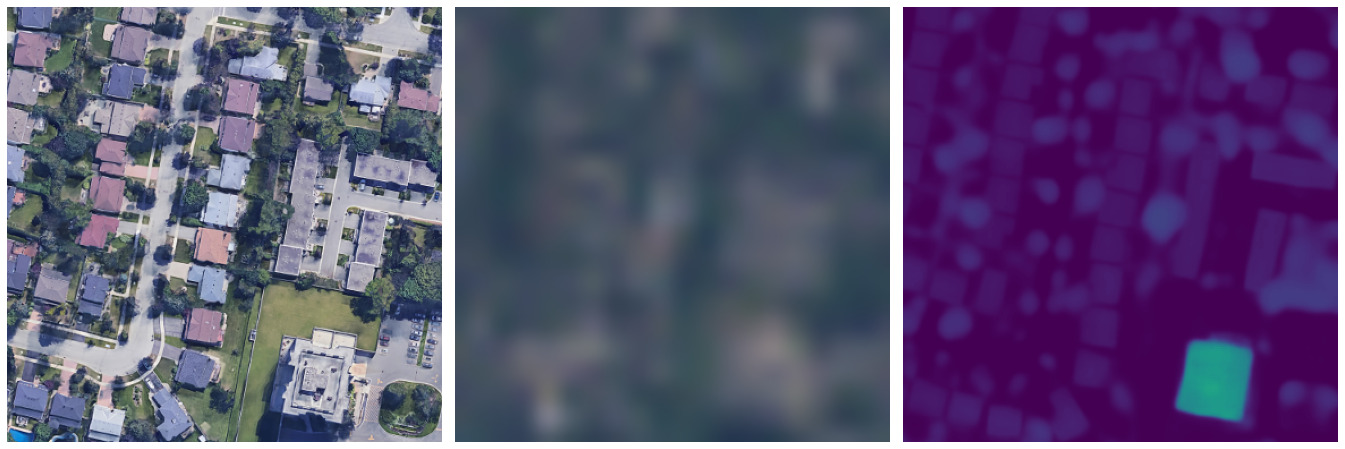}
    \caption{\label{fig:height-examples}Above-ground object height prediction (left: high resolution image for comparison, centre: Sentinel-2 RGB; right: predicted height mask). This method can predict building height with 1.5 meter mean absolute error.}
\end{figure}

In our work, we attempt to extend the limit of fine detail that can be recreated from a set of Sentinel-2 images, assessing the quality of building and road presence predictions made at 50 cm resolution. To do this, we use a teacher model with access to 50 cm resolution imagery to create training labels for a large dataset of worldwide imagery. We then train a student model to reconstruct these labels given only a stack of Sentinel-2 images from the corresponding places and times (see \Cref{fig:outputs}). We find that we are able to retain much -- though not all -- of the accuracy of the high-resolution teacher model: our Sentinel-2-based building segmentation has 79.0\% mIoU, compared to the high-resolution-based teacher performance of 85.5\% mIoU. We found that this accuracy level was equivalent to what could be achieved by a high-resolution model using a single frame of 4 m resolution imagery (see \Cref{tab:high-resolution-sensitivity}), though also noting through visual inspection that the segmentation quality sometimes far exceeds that (see \Cref{fig:building-road-examples} and \Cref{sec:appendix}).

\begin{figure}
  \centering
  \includegraphics[width=\textwidth]{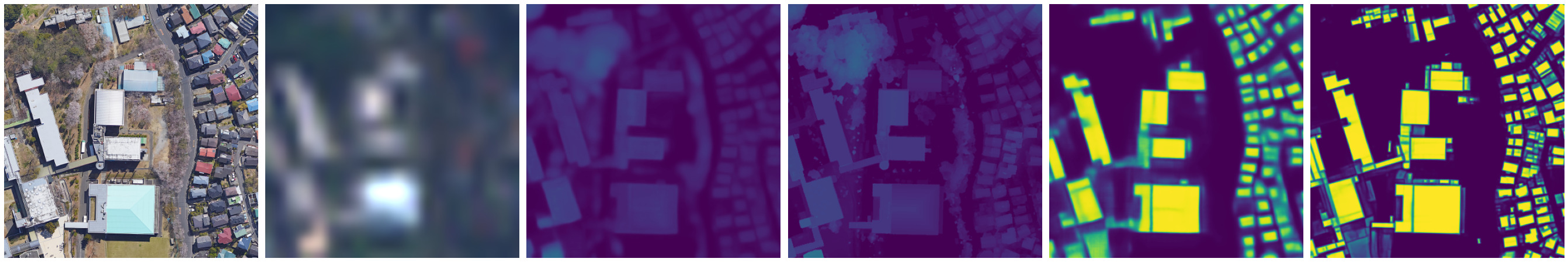}
  \includegraphics[width=\textwidth]{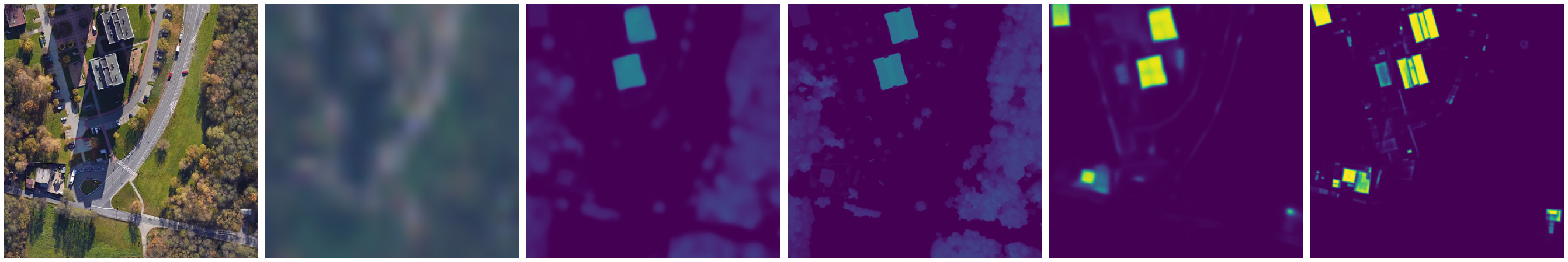}
  \includegraphics[width=\textwidth]{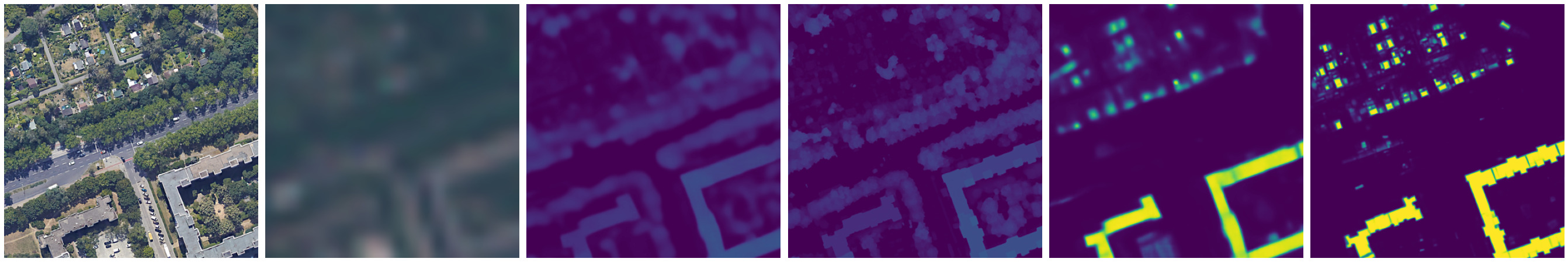}
  \includegraphics[width=\textwidth]{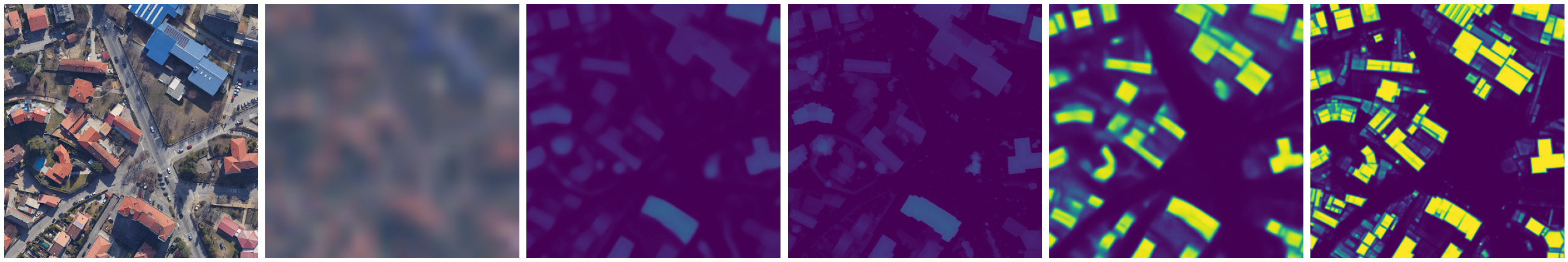}
  \includegraphics[width=\textwidth]{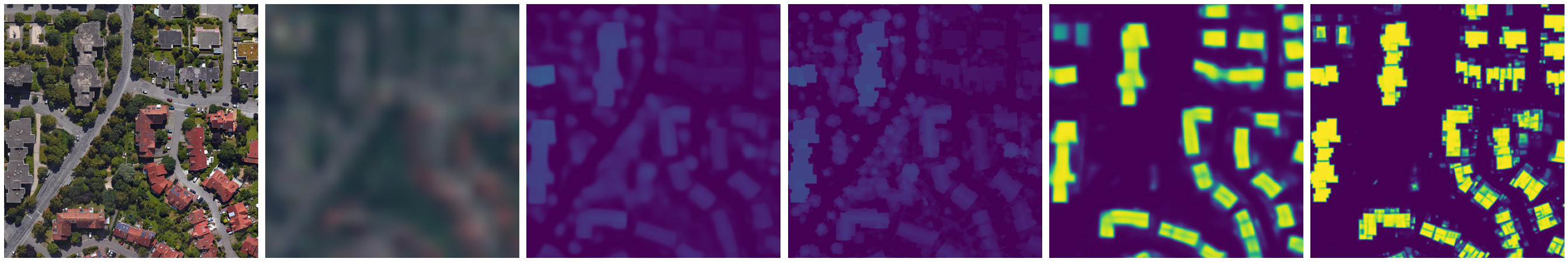}
  \includegraphics[width=\textwidth]{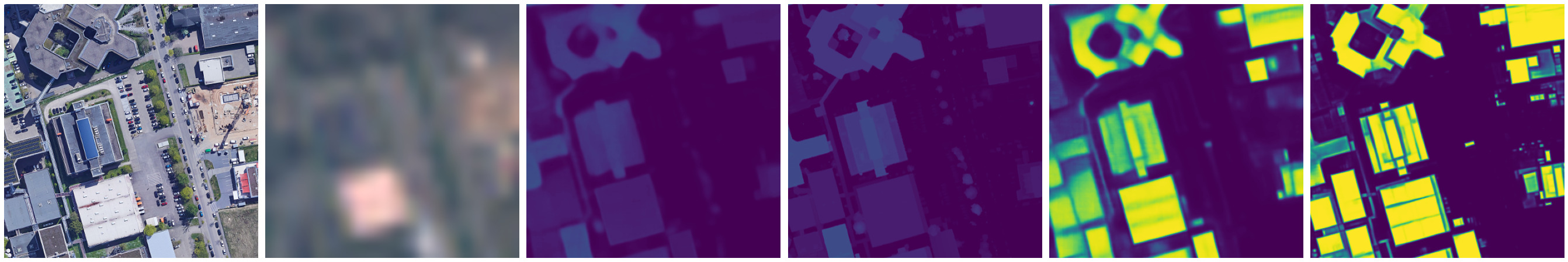}
  \includegraphics[width=\textwidth]{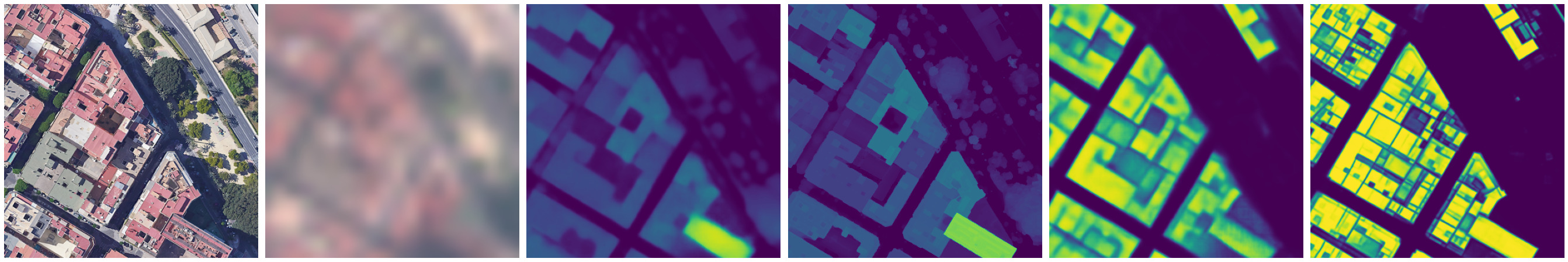}
    
  \begin{tabular}{*{6}{C{0.14\textwidth}}}
     \footnotesize 50 cm RGB & \footnotesize Sentinel-2 RGB & \footnotesize Height prediction & \footnotesize Height label & \footnotesize Building detection & \footnotesize Teacher building detection \\
  \end{tabular}

  \caption{\label{fig:appendix-height-examples}Examples of height prediction and building detection output, each covering an area of $192^2$ m$^2$.}

\end{figure}

We additionally describe experiments using the same model framework for two additional tasks: counting buildings and estimating building height. Building counts are estimated based on predicting the locations of building centroids. We can obtain $R^2 = 0.91$ compared to true building counts, again capturing much of the performance of the high-resolution teacher model ($R^2 = 0.95$). The predicted counts are surprisingly accurate even when buildings are very small relative to the raw Sentinel-2 resolution, or close together. Some examples are shown in Figures \ref{fig:centroid-count-examples} and \ref{fig:centroids_close}. Height prediction is done using per pixel above-ground object height labels during training. We find that the model is able to learn to predict building height with 1.5 meter mean absolute error. Examples are shown in Figures \ref{fig:height-examples} and \ref{fig:appendix-height-examples}.

This work extends the range of analysis tasks that can be carried out with freely available Sentinel-2 data. Buildings data produced by this model across Africa, South and Southeast Asia, and Latin America is now publicly available\footnote{\url{https://sites.research.google/gr/open-buildings/temporal/}}.

\section{Related work}
\label{sec:related}

Several researchers have noted the potential of using freely available but relatively low-resolution remote sensing imagery, such as Sentinel-2, to obtain insights about buildings and other features with previously unobtainable spatial and temporal scope. As well as experimental results, practical data is already being produced from such systems: for example, Sentinel-2 super-resolution from 10 m to 2.5 m has been used to create buildings data for 35 cities across China \cite{feng2023national}.

Super-resolution, the task of reconstructing a high-resolution image from one or more low-resolution images, has been widely applied to photographic images, commonly with generative models such as GANs. There is existing work on applying this type of model in remote sensing imagery \cite{pineda2020generative}, although there is a risk of increasing resolution at the expense of introducing spurious details. Certain characteristics of remote sensing imagery, and Sentinel-2 in particular, can be exploited to aid hallucination-free super-resolution. Alias and shift between sensor channels are undesirable for the purposes of visualization, but provide a useful signal for super-resolution \cite{nguyen2023role, beaulieu2018deep, pineda2020generative}. It is also possible to exploit aspects of the physical design of the Sentinel-2 satellites, to obtain 5 m super-resolved training data for specific areas with detector overlap \cite{nguyen2023l1bsr}.

Other results in the literature indicate that it is not necessary to explicitly model these alias and shift effects to obtain hallucination-free super-resolution, and that this can be learned directly from data with standard semantic segmentation models. U-Net is popular in remote sensing analysis and has been successfully applied in this setting. Simply up-sampling medium resolution imagery and trying to extract building detections gives increased effective resolution, which can then be used for building detection \cite{prexl2023potential}. Super-resolution followed by Mask-RCNN was used to detect buildings in locations across Japan \cite{chen2023large}. Another two stage model, SRBuildingSeg, carries out super-resolution followed by building segmentation \cite{zhang2021making}. A similar two stage setup based on U-Net was used to detect buildings using Sentinel-2 in Spain \cite{ayala2022pushing}.

Another principle useful for remote sensing super-resolution is to take multiple images to generate a single high-resolution output. Physical features such as buildings and roads change slowly in comparison to the Sentinel-2 revisit time of ~5 days, so an image stack is likely (though not guaranteed) to show the same scene. HighResNet \cite{deudon2020highres} is an architecture for fusing a temporal sequence of lower-resolution remote sensing images to predict a single, higher-resolution image. This works with a convolutional architecture to fuse the images together, and a loss function which can account for differences in alignment between the low-resolution input and high-resolution training labels. PIUNet (permutation invariance and uncertainty in multitemporal image super-resolution) is an architecture for multiple--image super-resolution, used to increase Proba-V resolution from 300 m to 100 m \cite{valsesia2021permutation}. Most of this existing work is based on fully convolutional architectures, however transformer-based models have also been used \cite{an2022tr}. The approach in our work is also to use multiple images to predict high-resolution features, although we do not have an intermediate super-resolution step and instead train end-to-end.

\section{Training setup}

\begin{figure}
  \centering
  \includegraphics[width=0.8\textwidth]{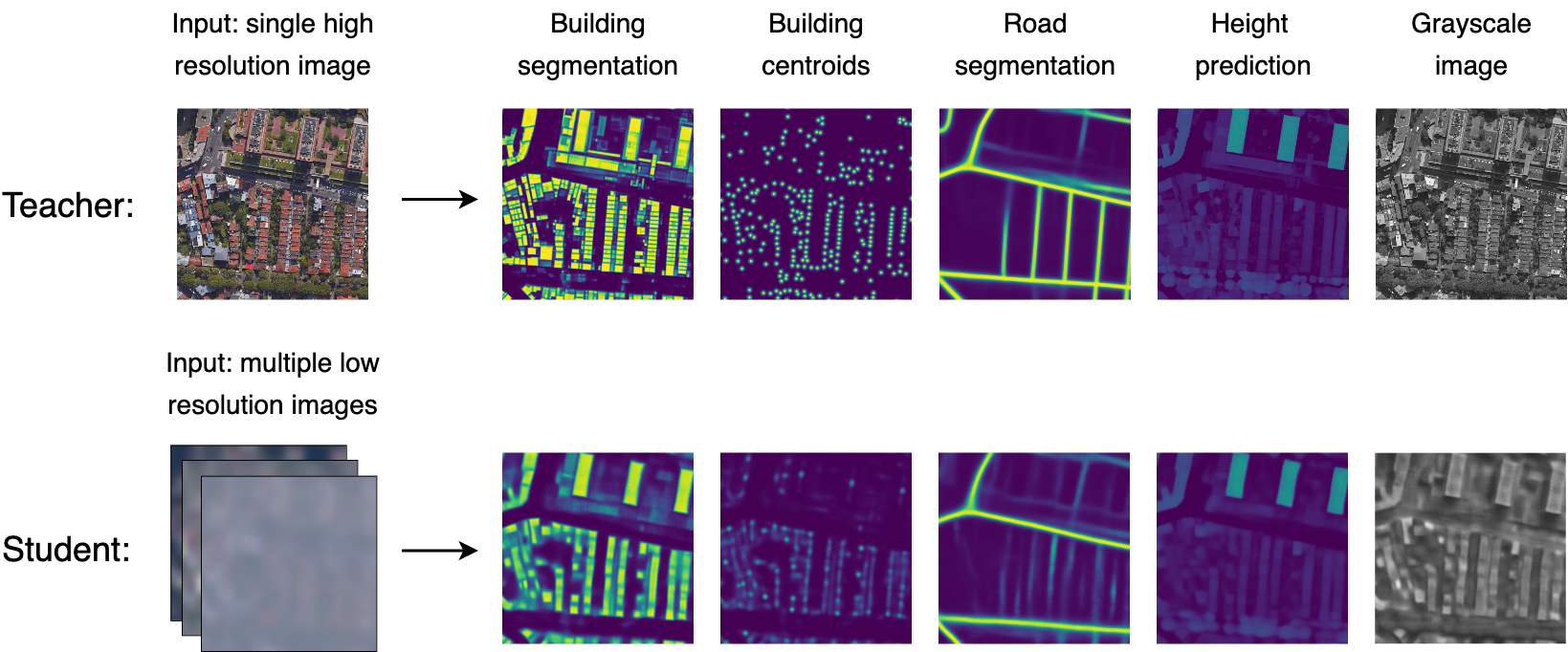} \vspace{.2cm}
  \caption{\label{fig:teacher-student}Teacher-student setup in this work. The student model is trained to reproduce the same outputs as a high-resolution model using 50 cm resolution imagery, but using only a stack of Sentinel-2 images at 10 m resolution.}
\end{figure}

We propose a method for semantic segmentation from a stack of low-resolution Sentinel-2 images at a much higher resolution than the input resolution. Our setup consists of a teacher model and a student model, with inputs and outputs as shown in \Cref{fig:teacher-student}. The teacher model is trained on high-resolution (50 cm) satellite imagery and outputs high-resolution semantic segmentation confidence masks for buildings, building centroids, and roads. The student model tries to mimic the output of the teacher model using only a stack of low-resolution imagery obtained from Sentinel-2 of the same location. 

Our method is an end-to-end super-resolution segmentation model, so that instead of first carrying out super-resolution on the image and then running semantic segmentation on the output, we make the model predict a high-resolution semantic segmentation mask directly from low-resolution inputs.

The input to the model is a set of low-resolution Sentinel-2 frames $\mathrm{LR}_{i,t}, \in \mathbb{R}^{H \times W \times C}$, arranged in a stack of time frames from $t = 1,\ldots,32$ (where $H \times W$ are spatial dimensions and $C$ is the number of input channels). The input channels include both imagery and metadata, as described in \Cref{sec:dataset}.   The output of the model is $\mathrm{SR}_i \in \mathbb{R}^{\gamma H \times  \gamma W \times C^{\prime}}$, where $\gamma$ is the upscaling factor and $C^{\prime}$ is the number of output channels. We denote the label by $\mathrm{HR}_i \in \mathbb{R}^{(\gamma H + \Delta y_{max}) \times  (\gamma W + \Delta x_{max}) \times C^{\prime}}$, where  $\Delta x_{max}$ and $\Delta y_{max}$ are margins corresponding to the maximum allowed translation in $x$ and $y$ direction respectively (\Cref{sec:loss_function}). 

Each output channel can potentially represent a separate task. We train models with up to five output channels: building semantic segmentation, roads semantic segmentation, building centroids, building heights and super-resolution grayscale image, as shown in \Cref{fig:teacher-student}. We use the building centroid predictions with one extra step to calculate building counts. We predict a super-resolution grayscale image as one of the output channels as this helps with image registration during training and evaluation.

At a high level, our model employs an encoder and a decoder. The encoder encodes each low-resolution image independently. The decoder takes a fused representation of these encodings and applies successive upsampling to it to output at target resolution. The model architecture is described in more detail in \Cref{sec:model}. 

\section{Dataset}
\label{sec:dataset}

\begin{figure}
    \centering
    \includegraphics[width=0.75\textwidth]{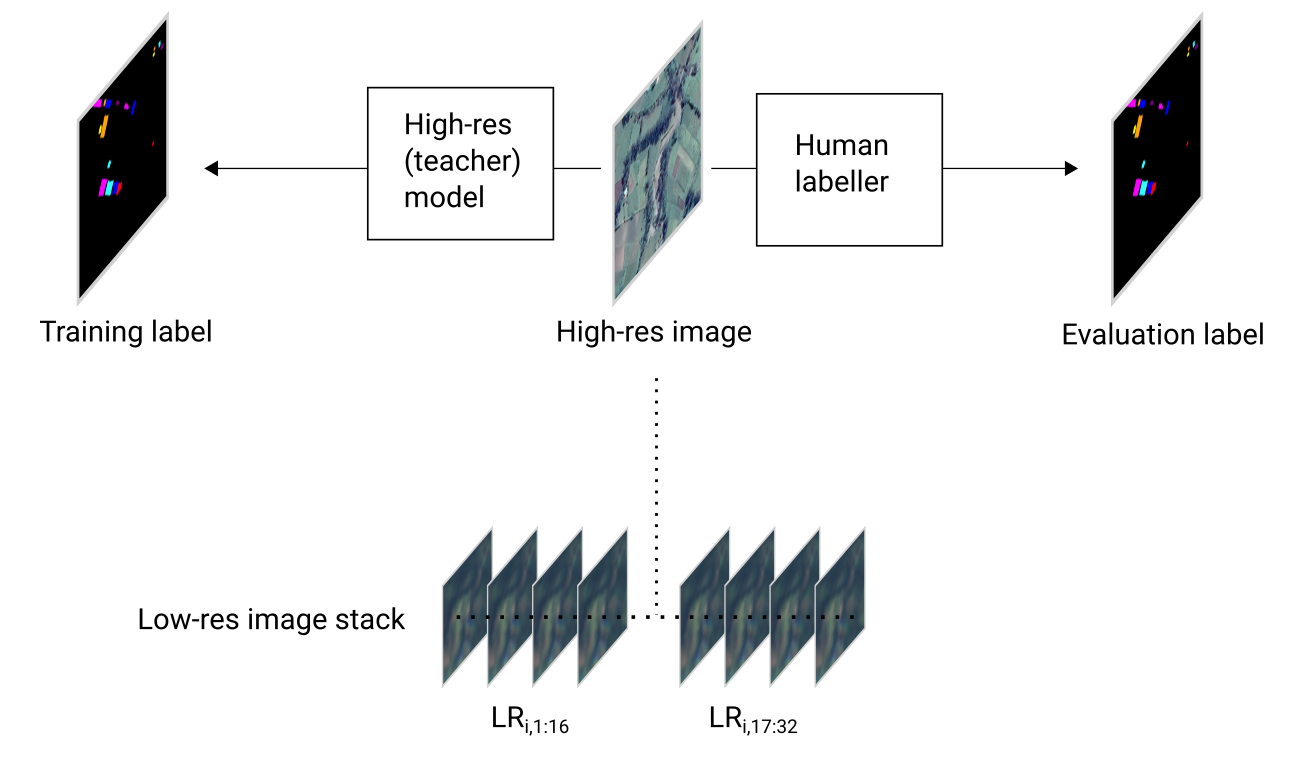}
    \caption{\label{fig:temporal-stack}Generation of training and evaluation data for the low-res student model. The low-resolution stack is constructed so that the time of the high-resolution image falls between the 16th and 17th frames.}
\end{figure}

\begin{figure}
    \centering
    \includegraphics[width=0.5\textwidth]{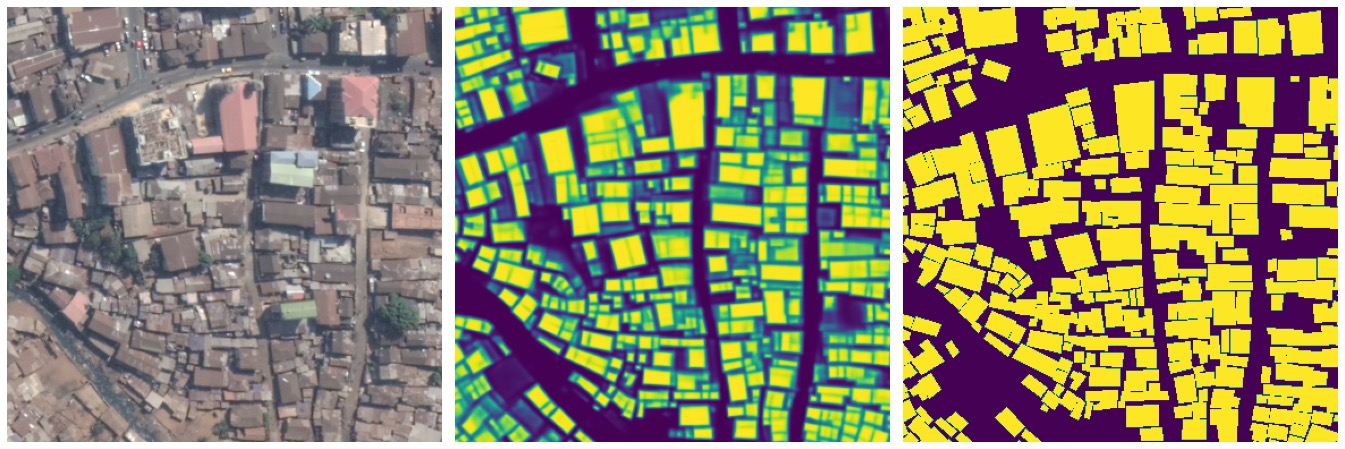}
    \caption{\label{fig:buildings_teacher_vs_human}Comparison of the model-derived teacher label (centre) and human-derived label (right) for one evaluation example.}
\end{figure}

The dataset for training and evaluation consists of low-resolution image stacks and high-resolution label pairs $\left\{\mathrm{LR}_{i,t=1:32}, \mathrm{HR}_i\right\}$. To generate labels for the temporal stack we fetch Sentinel-2 imagery stacks at locations where we also have high-resolution imagery available. We fetch the stack of Sentinel-2 imagery such that the high-resolution image corresponds to the middle of the low-resolution stack, i.e. between 16th and 17th time frame in a 32 frame stack (see \Cref{fig:temporal-stack}). Labels for the primary tasks of buildings and roads semantic segmentation are generated as follows: labels for the training split are per pixel building and roads presence confidence scores output by the teacher model, whereas labels for the evaluation split are binary segmentation masks obtained from human drawn polygons of buildings and roads on high-resolution images (see \Cref{fig:buildings_teacher_vs_human}). Since we generate training labels using a teacher model we can generate a large number of training examples, only limited by the amount of high-resolution imagery available and computational constraints.

The training split consists of two datasets: first one with about 10 million examples sampled globally where unorthorectified high-resolution satellite imagery was available and second one with 2.5 million examples sampled where high-resolution orthorectified imagery with detailed DSM (Digital Surface Model) and DTM (Digital Terrain Model) were available i.e. mostly in North America, Europe and Japan. For the first dataset, the examples are sampled randomly at uniform first and then subsampled by building presence using the high-resolution teacher model so that approximately 90\% of the examples have buildings, therefore as a consequence the distribution is roughly correlated with population density. The geographical distribution of examples is shown in \Cref{fig:dataset_geo_distribution}.

There are four validation datasets with human-derived building instance labels: "Africa" with 1158 examples, "Africa hard" with 945 examples, "South-Southeast Asia" with 1620 examples, and "Latin America" with 1631 examples. Furthermore, a separate validation dataset with above-ground object height labels (see \Cref{sec:height_labels}) but no human-derived building instance labels that consists of 1029 examples. The Africa dataset has examples which are all within the continent and chosen to include a range of urban and rural examples of different densities, as well as settings of humanitarian significance such as refugee settlements. "Africa hard" contains examples sampled from Africa where earlier high-resolution models had poor performance. The South-Southeast Asia and Latin America validation splits consist of examples where 75\% were sampled randomly across the regions and the other 25\% were sampled randomly while making sure environments with diverse building counts, building sizes and total building area were well represented. Training examples close to validation examples (within some radius) are discarded. For simplicity we do not perform any sampling based on roads. Often roads tend to be present in the vicinity of buildings, but this is not always true.

Sentinel-2 imagery is fetched from the Harmonized Sentinel-2 Top-Of-Atmosphere (TOA) Level-1C collection \cite{sentinel2ee} on Earth Engine. Each Sentinel-2 timeframe consists of 13 bands, each upsampled to 4 meters per pixel regardless of their native resolution. Additionally, we add metadata for each frame, as described in \Cref{sec:s2_data_proc} in more detail.

\subsection{Teacher labels}

\begin{figure}
    \centering
     \begin{subfigure}[b]{0.45\textwidth}
         \centering
         \includegraphics[height=2.5cm]{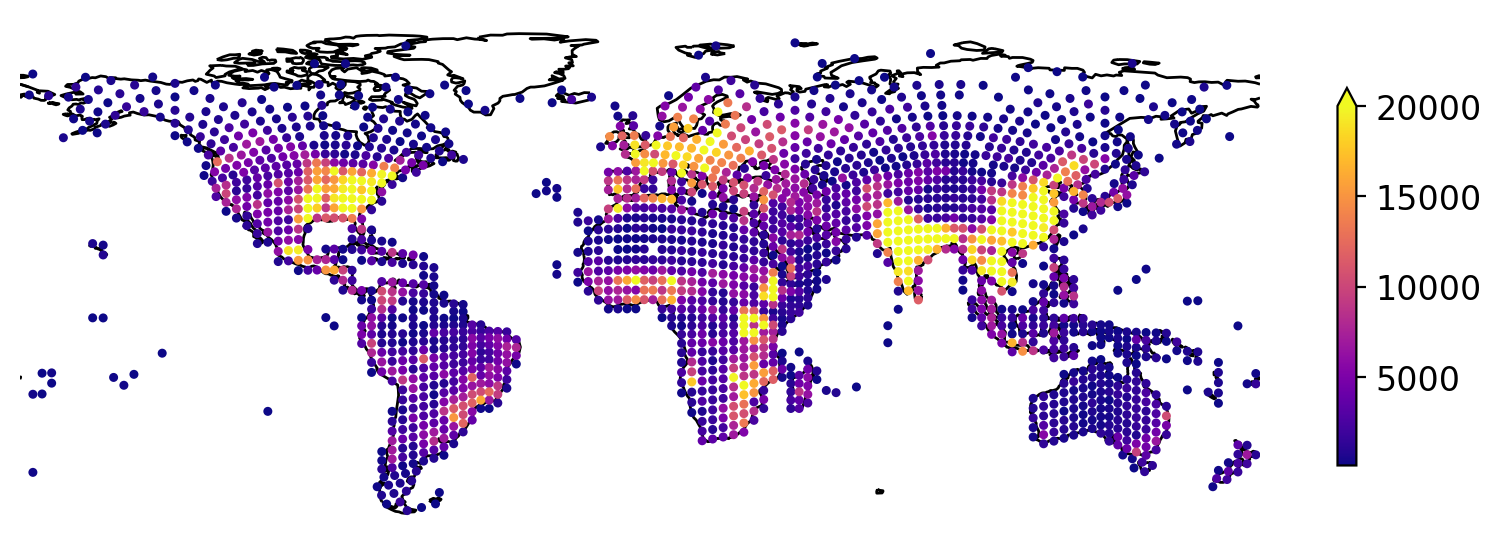}
         \caption{\label{fig:teacher_dataset_geo_dist}Teacher training: human labels (400K)}
     \end{subfigure}
     \hfill
     \begin{subfigure}[b]{0.49\textwidth}
         \centering
         \includegraphics[height=2.5cm]{figures/geo_distribution_student_unorthorectified_train.png}
         \caption{\label{fig:student_train_unortho_geo_dist}Student training: teacher labels (10M)}
     \end{subfigure}
     \hfill
     \begin{subfigure}[b]{0.53\textwidth}
         \centering
         \includegraphics[height=2.5cm]{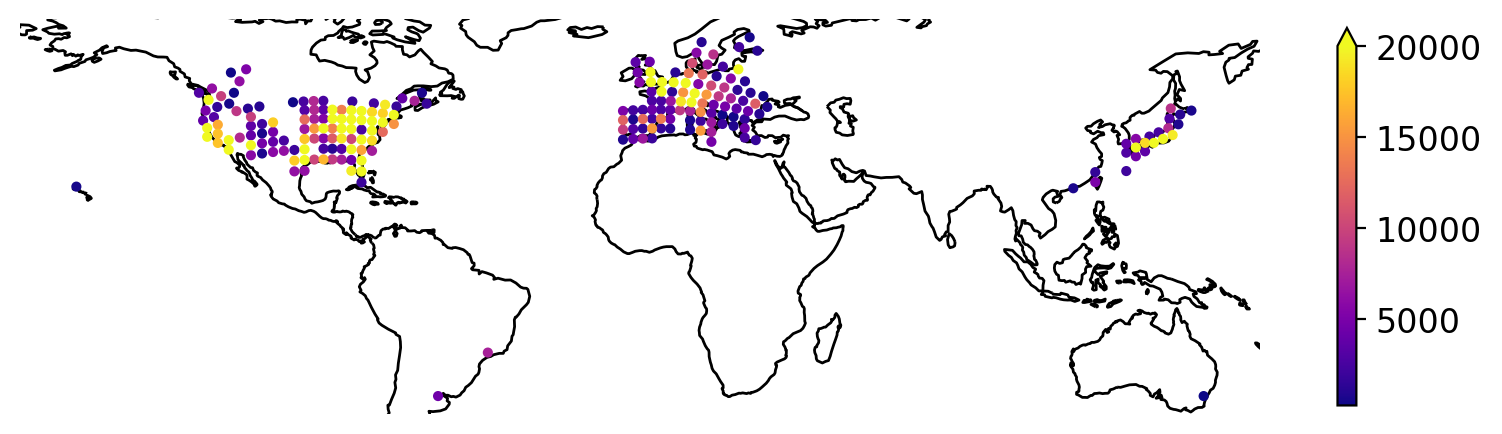}
         \caption{\label{fig:student_train_ortho_geo_dist}Student training: heights and teacher labels (2.5M)}
     \end{subfigure}
     \hfill
     \begin{subfigure}[b]{0.46\textwidth}
         \centering
         \includegraphics[height=2.5cm]{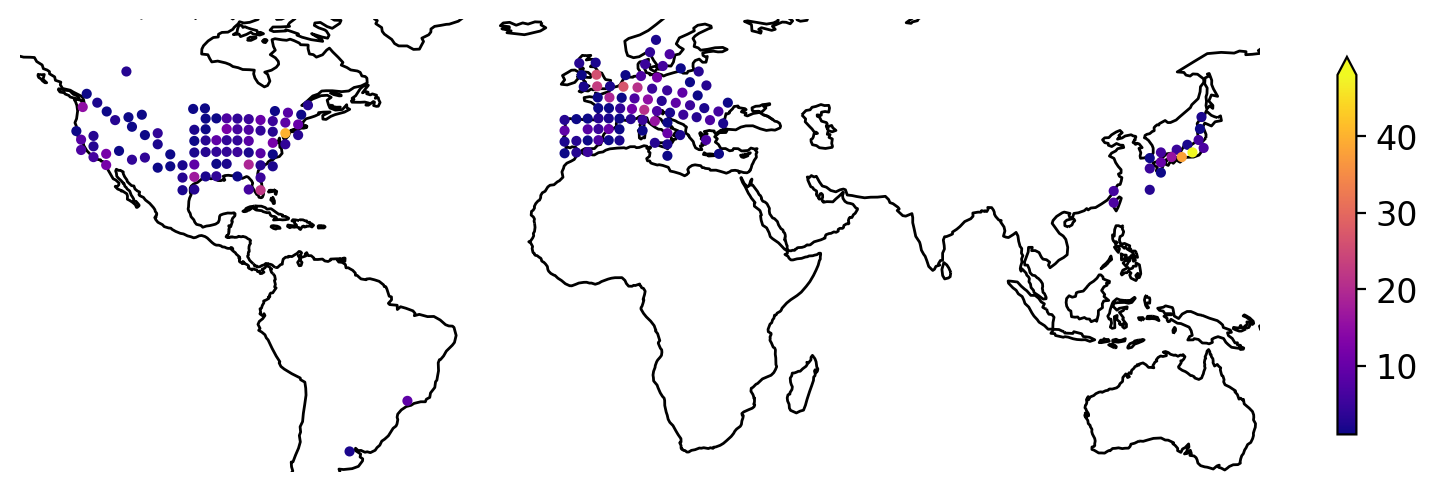}
         \caption{\label{fig:student_val_ortho_geo_dist}Validation: heights and teacher labels (1K)}
     \end{subfigure}
     \hfill
     \begin{subfigure}[b]{0.2\textwidth}
         \centering
         \includegraphics[height=2.5cm]{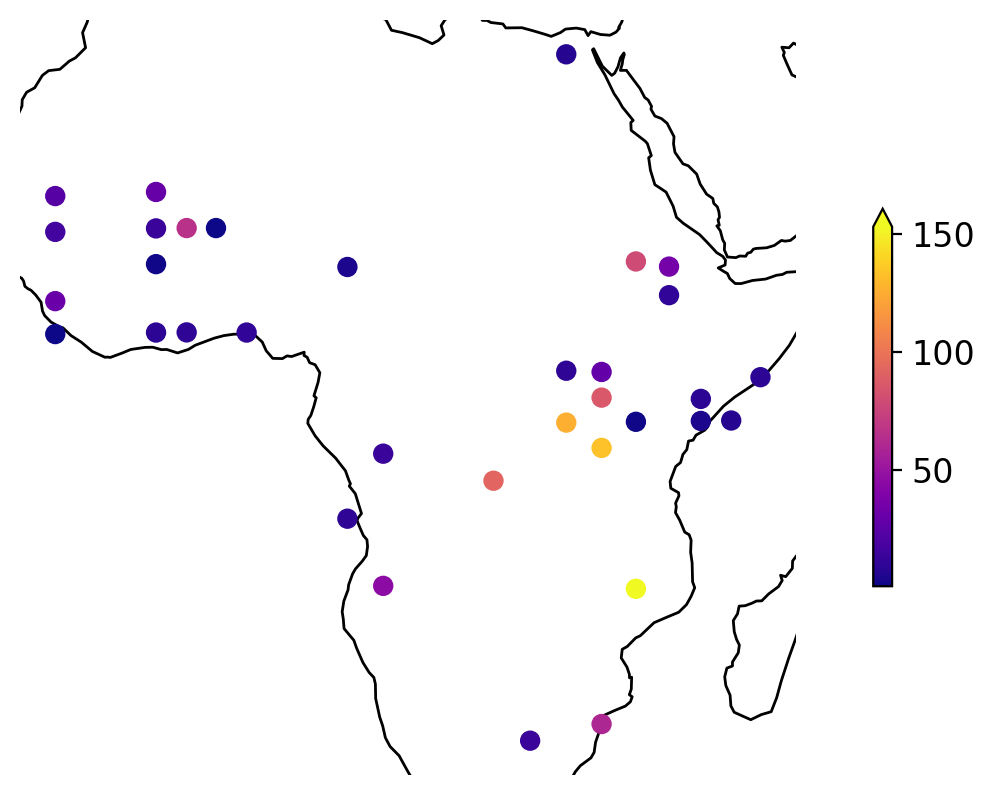}
         \caption{\label{fig:student_val_africa_geo_dist}Val.: Africa (1.2K)}
     \end{subfigure}
     \hfill
     \begin{subfigure}[b]{0.22\textwidth}
         \centering
         \includegraphics[height=2.5cm]{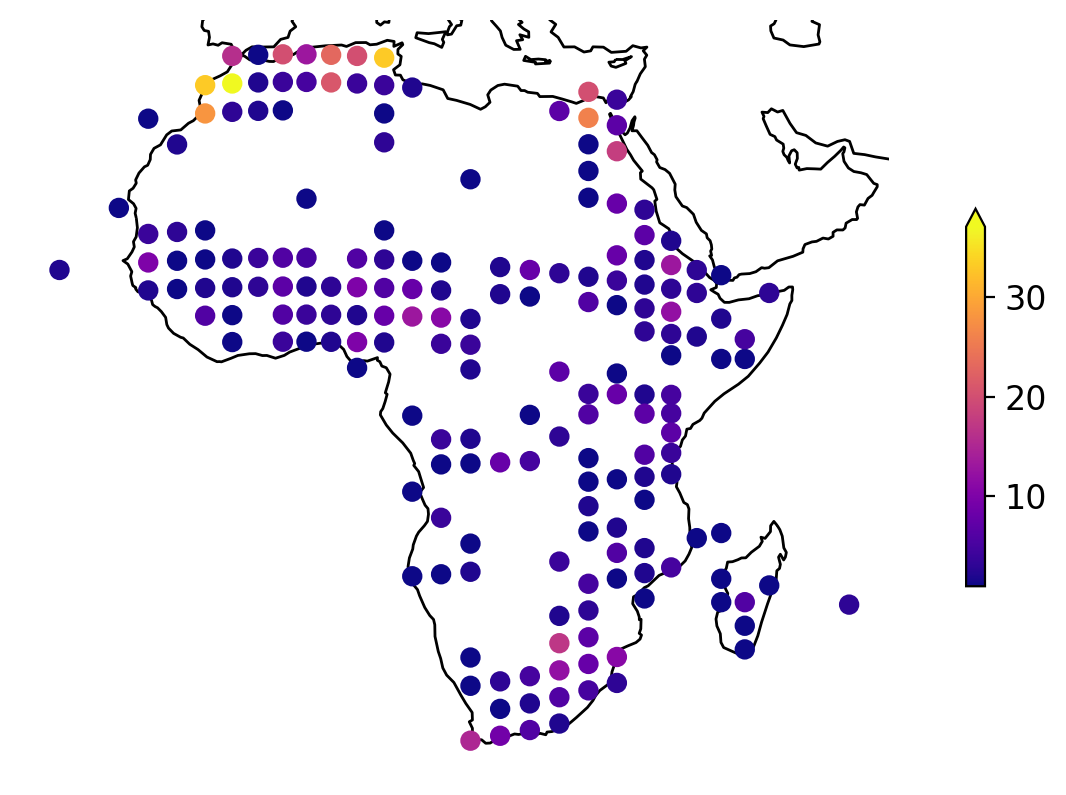}
         \caption{Val.: Africa hard (1K)}
     \end{subfigure}
     \hfill
     \begin{subfigure}[b]{0.3\textwidth}
         \centering
         \includegraphics[height=2.5cm]{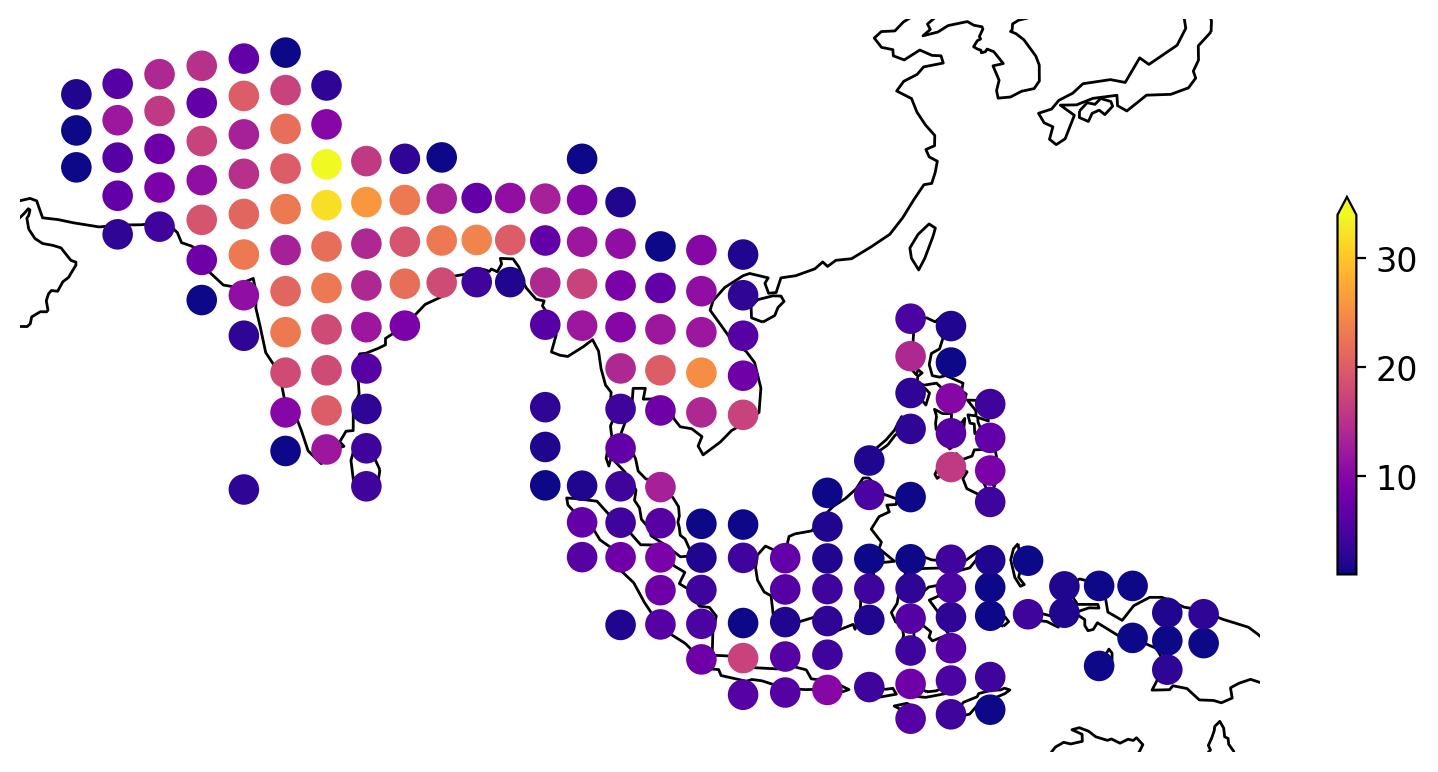}
         \caption{Val.: South-Southeast Asia (1.6K)}
     \end{subfigure}
     \hfill
     \begin{subfigure}[b]{0.24\textwidth}
         \centering
         \includegraphics[height=2.5cm]{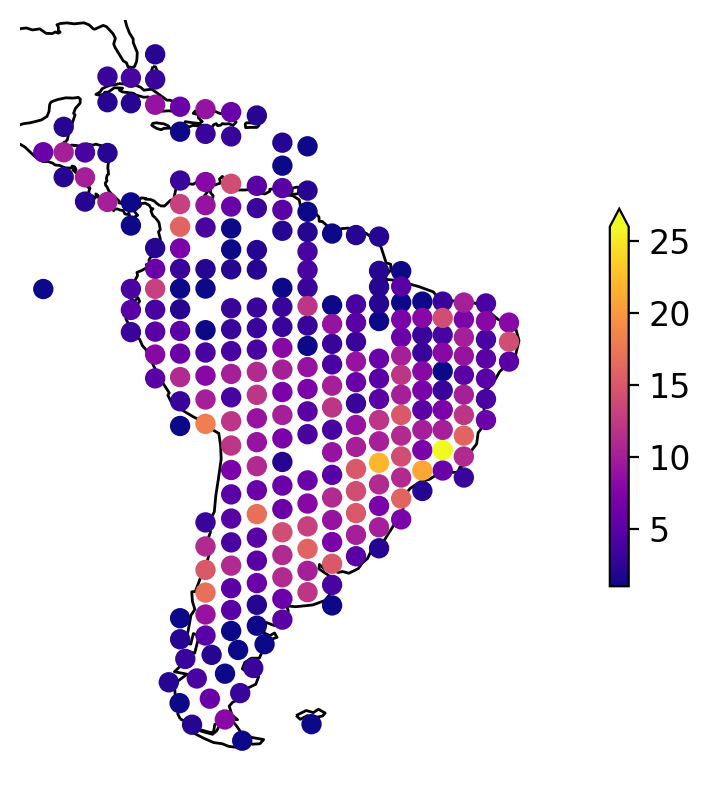}
         \caption{\label{fig:student_val_scac_geo_dist}Val.: Latin America (1.6K)}
     \end{subfigure}
    \caption{\label{fig:dataset_geo_distribution}Geographical distribution of all datasets used in this work. The teacher model was trained on the human-labelled high-resolution unorthorectified dataset in (\subref{fig:teacher_dataset_geo_dist}). The student model was trained on teacher-labelled unorthorectified dataset (\subref{fig:student_train_unortho_geo_dist}) and teacher-labelled orthorectified dataset with height labels (\subref{fig:student_train_ortho_geo_dist}). Furthermore, the student model was evaluated on datasets (\subref{fig:student_val_ortho_geo_dist})-(\subref{fig:student_val_scac_geo_dist}). The datasets (\subref{fig:student_val_africa_geo_dist})-(\subref{fig:student_val_scac_geo_dist}) are unorthorectified, human-labelled, and without height labels. Counts per level 5 S2 cell. The total number of examples is provided in parentheses.}
\end{figure}

The teacher model which is used to label the large worldwide training set is the same as used to generate the Open Buildings dataset, trained along the lines described in \cite{openbuildings}, with a few refinements for improved accuracy: these include the collection of extra training data in areas where earlier models had poor performance, ensembling of models trained on different partitions of the training data (distilled back to a single model) and use of the HRNet \cite{hrnet} architecture instead of U-Net. The geographical distribution of human-labelled training data for buildings is shown in Figure \ref{fig:dataset_geo_distribution}, reflecting the current Open Buildings coverage across Africa, South and Southeast Asia, and Latin America. Training is done in a similar way for road detection as for building detection.

\subsection{Above-ground object height labels}
\label{sec:height_labels}

\begin{figure}
    \centering
    \includegraphics[width=0.5\textwidth]{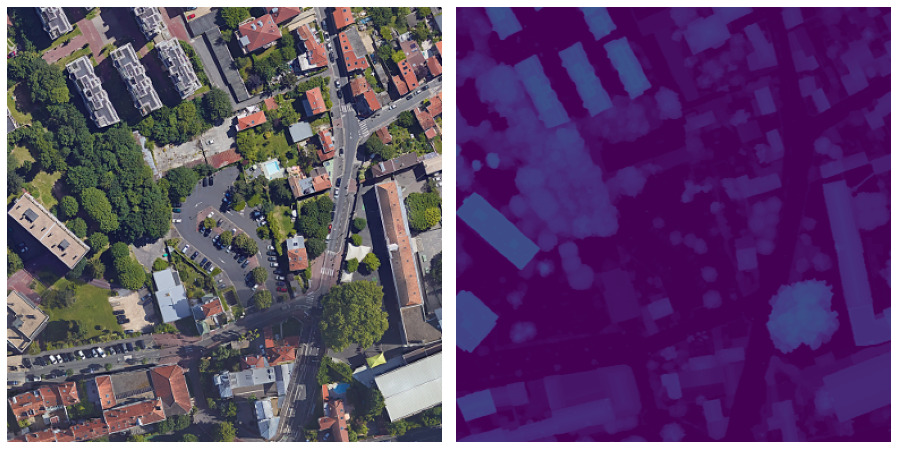}
    \caption{\label{fig:height_label_example}Above-ground object height label example.}
\end{figure}

Approximately 20\% of the examples in the datasets were sampled in locations where we had access to a detailed DSM (Digital Surface Model) and DTM (Digital Terrain Model) matching time-wise the orthorectified high-resolution image. By subtracting the DTM label from the DSM label we obtained above-ground object height labels that for a given pixel denote the height in meters of objects such as buildings, bridges and trees. Before the height labels were generated the DTM label was smoothed out to a consistent value for all pixels within a single building instance. As a consequence of how labels are generated the model is, among others, able to predict vegetation height. However, in this work we only focus on the building height prediction aspect, therefore for height prediction evaluation we ignore the non-building pixels using the teacher labels (human-derived building labels were unavailable). An example of a above-ground object height label is shown in \Cref{fig:height_label_example}.

\subsection{Registration}

Neither low-resolution Sentinel-2 images nor the high-resolution image are registered to any reference frame, and as such, both the input imagery stack as well as the labels are potentially misaligned relative to each other. Given the resolution of Sentinel-2 imagery, misalignments of a few pixels in image space amount to tens of meters on the ground. We rely on the model being able to implicitly align the low-resolution input frames. However, model output can still be misaligned relative to the labels, and in order to not penalise the model for that we manually align labels to model output during loss computation, as described in \Cref{sec:loss_function}. We also do similar alignment in evaluation.

\subsection{Clouds}

To filter cloudy images we discard Sentinel-2 images that have one or more pixels in the opaque cloud mask set. An opaque cloud mask is derived from the QA60 band in the Sentinel-2 TOA collection (10th bit). QA60 band however is inaccurate and does not always capture all clouds (see example in \Cref{fig:inaccurate_cloud_mask}). We retain Sentinel-2 images even if QA60 indicates they have cirrus clouds, because often these images still looked useful. Moreover, cloudy high-resolution images are especially problematic since they introduce noise into the labels. Therefore both the training and evaluation datasets are filtered to have cloud-free high-resolution imagery. During inference, we only use the QA60 band to filter out timeframes with opaque clouds.

\subsection{Orthorectification}
\label{sec:orthorectification}

\begin{figure}
\centering
\begin{minipage}{.45\textwidth}
  \centering
  \includegraphics[height=.15\textheight]{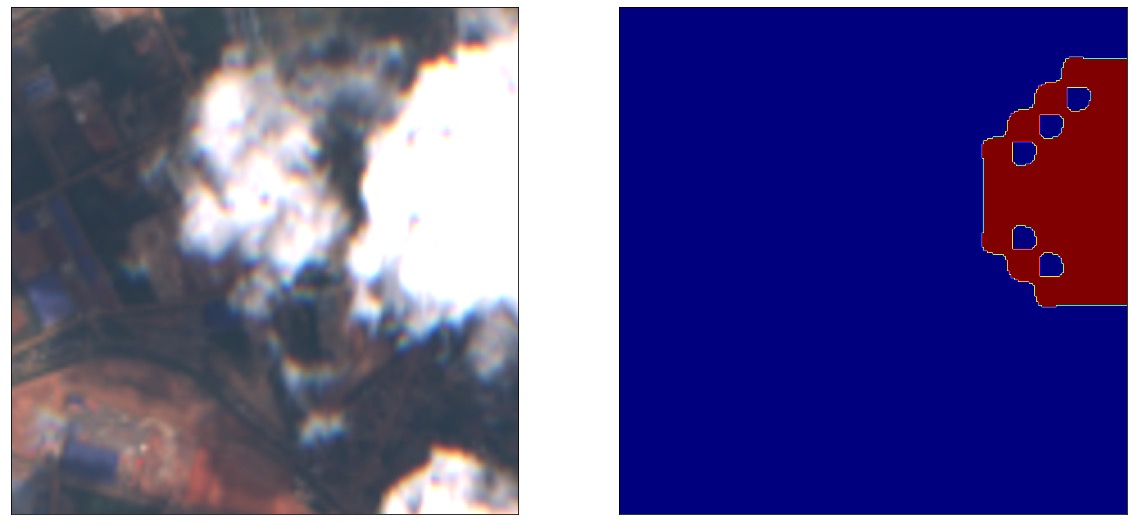}
  \captionof{figure}{\label{fig:inaccurate_cloud_mask}Inaccurate QA60 band cloud mask.}
\end{minipage}
\qquad
\begin{minipage}{.45\textwidth}
  \centering
  \includegraphics[height=.15\textheight]{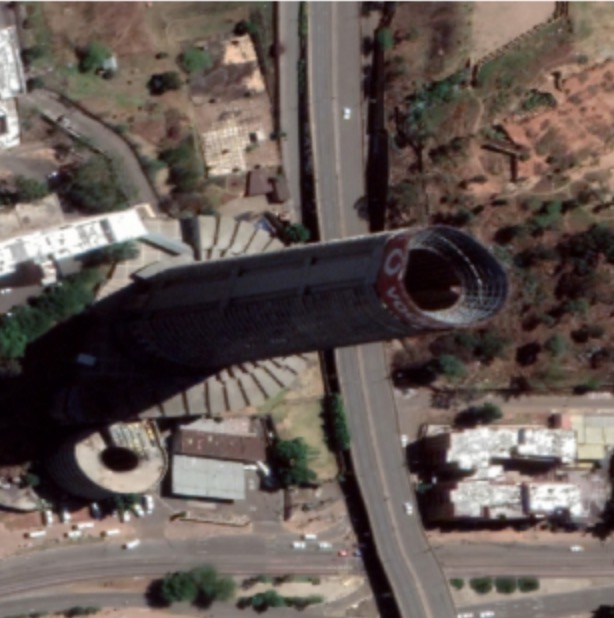}
  \captionof{figure}{\label{fig:unorthorectified_imagery}Non-orthorectified image: tall structure.}
\end{minipage}
\end{figure}

Label transfer from high resolution to low resolution works well only if both the satellites have the same viewing angle or both the imagery collections are orthorectified. 80\% of the high-resolution imagery we use is however not orthorectified (see example in \Cref{fig:unorthorectified_imagery}) and as such, labels generated using such imagery do not always align with Sentinel-2 imagery. Particularly problematic are areas with tall buildings where building roofs (which is what our teacher model is trained to detect) appear in different positions depending on the satellite viewing angle. Orthorectification of satellite imagery however requires an accurate elevation model -- something that is not available globally (see \Cref{fig:student_train_ortho_geo_dist}). In the absence of orthorectification, we pass the high-resolution satellite image incidence angle metadata on input to the model. See \Cref{sec:experiments/high_res_incidence_angle} for more details.

\subsection{Sentinel-2 data processing}
\label{sec:s2_data_proc}

Due to the way imagery assets are tiled in the 
\href{https://developers.google.com/earth-engine/datasets/catalog/COPERNICUS_S2_HARMONIZED}{Harmonized Sentinel-2 Top-Of-Atmosphere (TOA) Level-1C collection} on Earth Engine, the imagery stack fetched for a given location can potentially contain duplicate imagery \cite{link_datatake}. We therefore perform deduplication of the imagery stack as follows: we group images by datatake ids and for each datatake id we take the highest processing baseline.  Datatakes correspond to a swath of imagery taken by the Sentinel-2 satellite and often cover a very large surface area of the earth (up to 15,000 square kilometers). The processing baseline is effectively a set of configurations used to post process raw imagery acquired from the satellites and are updated regularly by the European Space Agency (ESA).

We feed in metadata associated with each frame to the model using a simple approach where scalar value for each metadata item is broadcast to image spatial dimensions and then appended to the image in the channel dimension. As such, each scalar metadata value adds an additional channel to the input. The following metadata are used for each frame: normalized time relative to the 17th time frame, mean incidence azimuth angle, mean incidence zenith angle, mean solar azimuth angle, mean solar zenith angle, latitude, longitude, mean high-resolution satellite image incidence azimuth angle and zenith angle (the last 4 are the same for each frame). The time of image acquisition is normalised by dividing the time duration relative to the 17th frame by ten years in seconds, and scale the other features to the [0, 1] range. We do not perform any data augmentation other than random cropping, which we found is sufficient to prevent overfitting due to the very large dataset size. Mean high-resolution satellite image incidence azimuth angle and zenith angle are set to $0^\circ$ during inference.

To account for the fact that, in certain cloudy parts of the world it is not always possible to acquire a stack of 32 Sentinel-2 images in a given time span, during training we randomly truncate and pad the image stack on both ends. More formally, for both halves of the stack, with some probability $p$ we generate a padding of length $l$ distributed uniformly in $[1, 16]$. This helps to make the model robust to missing frames at inference time.

\section{Model}
\label{sec:model}

\begin{figure} 
    \centering
    \includegraphics[width=\textwidth]{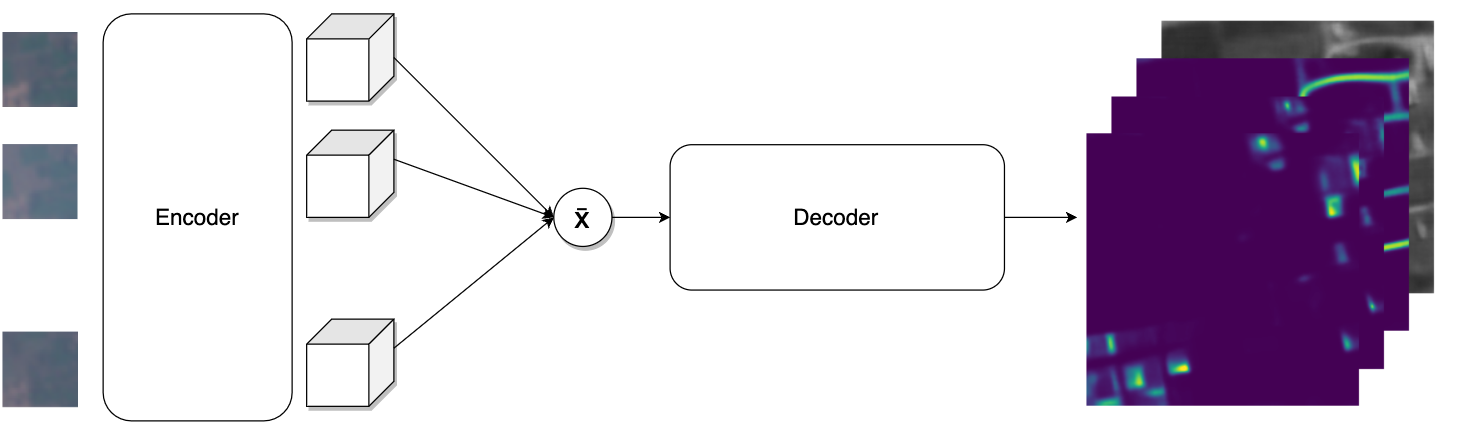}
    \caption{\label{fig:overall_arch}Overall model architecture. Each of the input images is encoded separately using a shared encoder. Encodings are then averaged together (which collapses the time dimension) and passed on to the decoder that upsamples the fused encoding by a factor of 8.}
\end{figure}

At a high level our student model consists of a shared encoder module that encodes each LR frame separately, a simple mean based encoding fusion and a decoder module that up-samples the fused encoding (\Cref{fig:overall_arch}). Below we describe encoder and decoder modules in detail.

\subsection{Encoder}
\label{sec:model_encoder}

\begin{figure}
    \centering
    \includegraphics[width=0.9\textwidth]{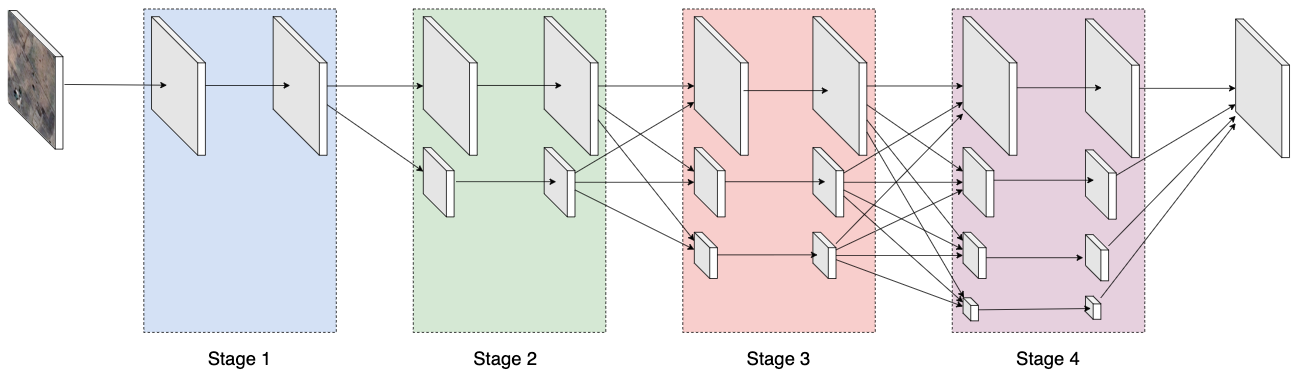}
    \caption{\label{fig:hrnet-single-frame}HRNet architecture, as it would be applied to a single frame of imagery. The network consists of 4 different stages and each stage consists of blocks that correspond to features at different resolutions. Each Sentinel-2 input has 13 image channels and 7 metadata channels.}
\end{figure}

We employ the HRNet \cite{hrnet} architecture (see \Cref{fig:hrnet-single-frame}) as an encoder without making any significant modifications, except for the root block. The original root block in HRNet downsamples the input image by a factor of 4, which we find to be too aggressive for LR input images that are already quite low resolution. To address this, we remove the 4x downsampling by reducing stride from 2 to 1 on the two 3x3 convolutions in the first block of HRNet. As a result, the encoder outputs features at the same spatial resolution as the input. We pre-train the HRNet encoder on ImageNet. To adapt filter weights of the first convolution trained on 3 channel RGB ImageNet input to more channels, for each filter we take the mean of the weights across channels and replicate the mean across the target channel dimensions (=13+9, corresponding to all bands in a Sentinel-2 image and all metadata we pass in).

\subsection{Cross-time information fusion}
\label{sec:cross_time_fusion}

\begin{figure}
    \centering
    \includegraphics[width=0.9\textwidth]{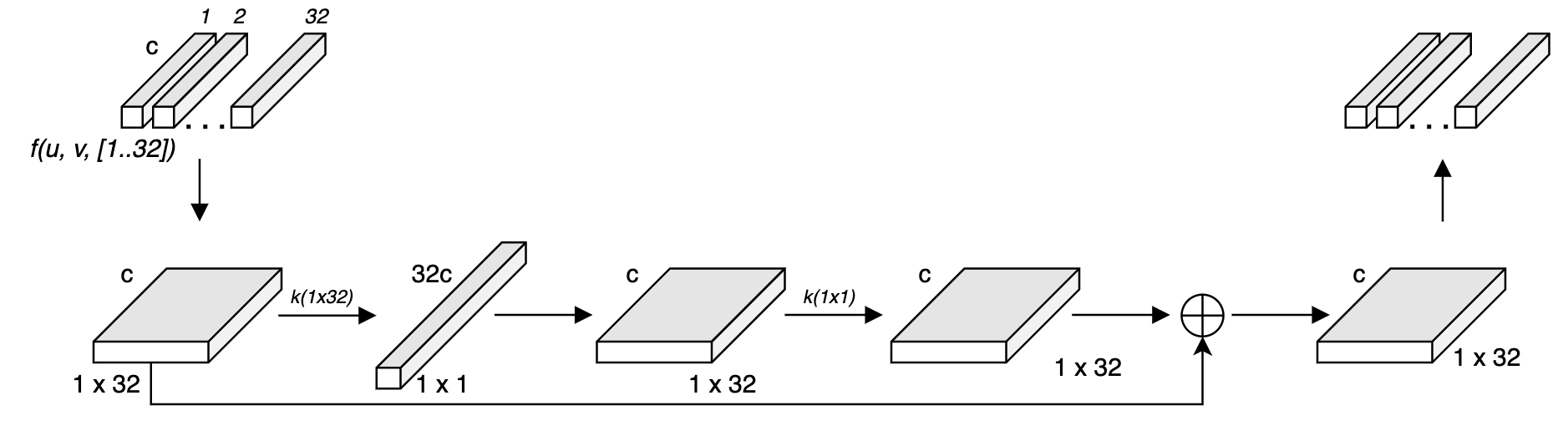}
    \caption{\label{fig:cross_time_fusion}cross-time fusion of features. For each stage and block (corresponding to features at various depths and resolutions) spatial features for pixel $(u, v)$  across all 32 time steps are fused using depth-wise convolution in a residual fashion.}
\end{figure}

In order to make the model learn not just spatial features but also the temporal relationship between spatial features we experimented with cross-time information fusion as features go through the various stages of HRNet encoder. For each stage and block (corresponding to features at various depths and resolutions), spatial features are fused across time using a depth-wise convolution in a residual fashion (\Cref{fig:cross_time_fusion}). 

For a given block and stage, features for pixel $(u, v)$  across all 32 time steps  are passed through a $1\times32$ depth-wise convolution (with depth multiplier $32$) to transform a $1\times32\times c$ feature tensor to $1\times1\times32c$. The tensor is then reshaped to $1\times32\times c$ before being passed to $1\times1$ point-wise convolution to obtain $1\times32\times c$ fused features.  We reshape the tensor before the point-wise convolution to reduce the number of convolution parameters. Otherwise, this setup is equivalent to a 1D depth-wise separable convolution across time.

Note that this fusion approach is much more efficient than 3D convolution or ConvLSTM in terms of the number of parameters. This is because our approach only fuses features across time for a single pixel, and using depth-wise convolution (with reshaping before point-wise convolution) helps to reduce the number of parameters even further.

The original features are then added to the fused features as a residual connection. More details about pairing schemes and experimental results are given in \Cref{sec:pairing_schemes}.

\subsection{Decoder}

\begin{figure}
    \centering
    \includegraphics[width=0.7\textwidth]{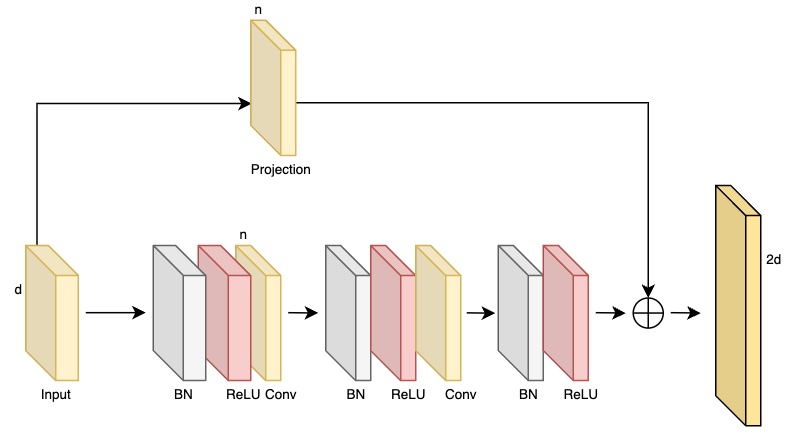}
    \caption{\label{fig:residual_decoder}Residual Decoder block with width $n$ that upsamples the input by a factor of 2.}
\end{figure}

The decoder module in our model consists of a number of decoder blocks that each upsample the input by a factor of 2. We use the same decoder block as in our previous technical report \cite{openbuildings} which consists of  x2 (batch normalization, ReLU, convolution) followed by another (batch normalization, ReLU), fusion with residual connection to the input and finally an upconvolution, as illustrated in \Cref{fig:residual_decoder}. We use transposed convolution or deconvolution for up convolution. The fused encoding is passed through 3 decoder blocks with widths 360, 180 and 90 respectively. Finally the upsampled features are passed through a (convolution, batch normalization, ReLU) block before being passed through a final convolution that outputs the desired number of output channels. There is one output channel for each task: building segmentation, road segmentation, building centroid detection, building height prediction and image super-resolution. Since the input Sentinel-2 frames are the same for each of these tasks, a single model can be trained in a multi task fashion to perform these tasks simultaneously. Apart from reducing cost and time, the different tasks could possibly serve as regularization for the other tasks to produce better results.

\subsection{Building counts via centroid prediction}
\label{sec:hi_res_teacher_model/extension_to_centroids}

\begin{figure}
    \centering
    \includegraphics[width=\textwidth]{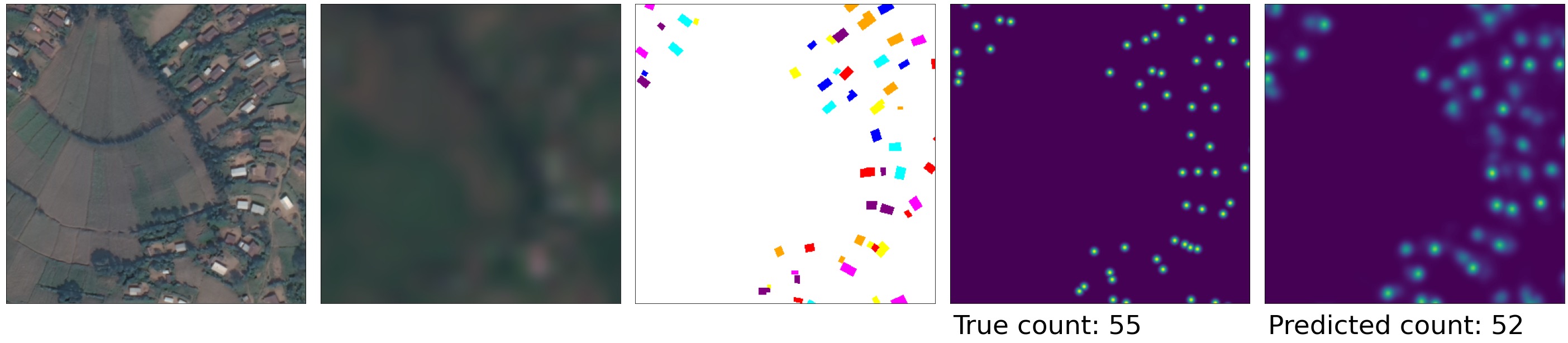}
    \includegraphics[width=\textwidth]{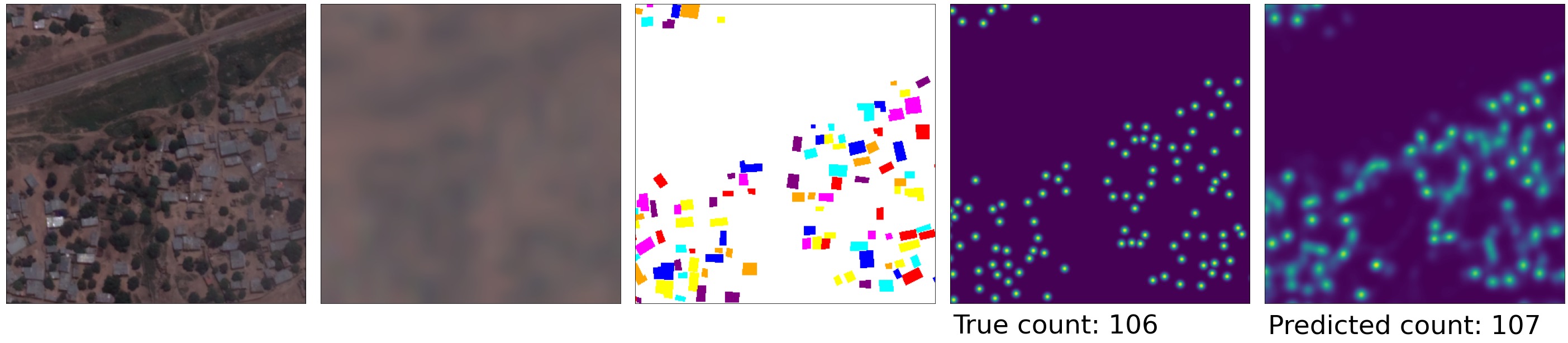}
    \caption{\label{fig:centroids_close}From the left to the right high-resolution image, the first images of the corresponding Sentinel-2 stack, the instance masks derived from teacher model, corresponding centroid labels, and model output.} 
\end{figure}

In some practical situations, people need building counts in an area rather than a segmentation mask of building presence. We investigated various ways of trying to estimate the count of buildings in an image tile from the Sentinel-2 image stack. The most effective method we found is based on first predicting the positions of building centroids, inspired by \cite{zhou2019objects} and other work on object counting e.g. for crowds. For the centroid detection task, the labels are Gaussian `splats', as shown in \Cref{fig:centroids_close}, where each splat is the same size regardless of the size of the building. This means that we expect a roughly constant response for each building, and we can derive an estimate count by simply summing the model output over the spatial dimensions and then dividing by a constant scaling factor.

We therefore derive the count without requiring any post-processing such as peak detection or non-maxima suppression. For a given tile $i$, the building count, $\hat{n}_i$ can be estimated as follows:

\begin{equation}
    \hat{n}_i = \dfrac{1}{K} \sum_{u,v} C_{u,v}^{i}\ ,
\end{equation}
where $C_{u,v}^{i}$ is model output at pixel $u,v$ (for the channel corresponding to centroid labels) and $K$ is the scaling factor. $K$ is computed as the average of per-sample scaling factors across many training set samples. The per-sample scaling factor is derived from the centroid labels sum and the true count for that sample.

The centroid detection task is incorporated into the overall model as one of the output channels. Building instances from the teacher model, above a certain score threshold, are used to generate centroid labels. See \Cref{fig:centroids_close} for example labels and model output. We detail some of the results obtained with this approach in \Cref{sec:experiments/extension_to_centroids}.

\section{Loss function}
\label{sec:loss_function}

We employ per pixel Kullback-Leibler Divergence (KLD) loss between the label and the prediction, defined as follows:

\begin{equation}
KLD(y_{i}, \hat{y}_{i}) = \left(y_{i} \log\left(\frac{y_{i}}{\hat{y}_{i}}\right) + \left(1 - y_{i}\right) \log\left(\frac{1-y_{i}}{1-\hat{y}_{i}}\right) + \epsilon \right)^\gamma\ ,
\label{eq:kld-loss}
\end{equation}
where $y_{i}$ and $\hat{y_{i}}$ are the label and prediction for pixel $i$ respectively and $\gamma$ is the focal term. Both $y_{i}$ and $\hat{y_{i}}$, originally $\in [0, 1]$, are clipped to $[\epsilon, 1-\epsilon]$, where $\epsilon = 1e-7$, to avoid division by zero errors. The focal term $\gamma$ varies the importance given to hard (misclassified) examples. Following a hyper-parameter sweep we set $\gamma$ to 0.25. We also add $\epsilon$ to KLD before exponentiation by $\gamma$ to guard against an undefined gradient. For examples where the above-ground object height label is missing we zero out the loss term for the height task.

Since the label and the model output could be misaligned, and in order to not penalize the model due to misalignment, we do an exhaustive neighborhood search based registration between the label and the prediction. For this we assume a translation model and find the $(\Delta x, \Delta y)$ translation that minimizes the mean squared error (MSE) between the label and prediction. The label is then shifted by $(\Delta x, \Delta y)$ to register it with the model output, before the loss in (\ref{eq:kld-loss}) is computed. This exhaustive neighborhood search is however computationally expensive. A possible alternative would be to also learn registration as proposed in \cite{deudon2020highres}, where a separate registration module takes in (prediction, label) pair as input and predicts a $k\times k$ kernel which when convolved with the model output aligns it with the label. 

\begin{figure}
    \centering
    \includegraphics[width=0.3\textwidth]{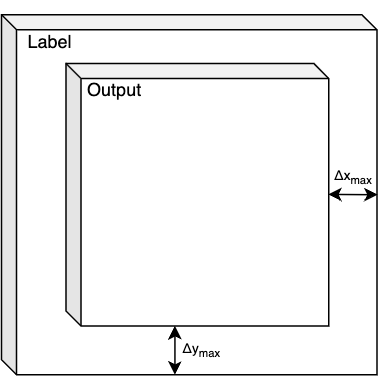}
    \caption{\label{fig:target_with_margin}Alignment at loss: Label with margin whose size is equal to the maximum allowed translation along a given image dimension.}
\end{figure}

We crop the label such that we keep a margin around it whose size is set to the maximum translation $(\Delta x_{max}, \Delta y_{max})$ allowed in a given direction (\Cref{fig:target_with_margin}). Once the label is shifted relative to the model output we crop the label to the output dimension so that both the label and model output have the same size.

To make alignment more robust, we also make the model predict a super-resolved grayscale image. The label for that task is the 50 cm image converted to grayscale. We use similar registration logic in both training and evaluation.

\section{Experiments}
\label{sec:experiments}

In our experiments, we report the mean Intersection over Union (mIoU) for a binary segmentation setup, where the foreground is either buildings or roads. Instead of using a fixed threshold (such as 0.5) to convert the per-pixel confidence values output by the model into a segmentation mask, we use a threshold in the range $[0, 1]$ and a dilation kernel size that maximizes the mIoU. We then report the maximum mIoU obtained. For the auxiliary task of tile level building count, we report mean absolute error (MAE) and the coefficient of determination, R$^2$.

The numbers between different tables are not necessarily comparable due to slight differences in configurations. Our evaluations focus on the building detection task and we do not report metrics for the road detection and image super-resolution tasks. Unless otherwise noted evaluation numbers are reported for the Africa validation dataset.

\subsection{Training Details}

Our models are trained with the Adam optimizer \cite{kingma2014adam} using a constant learning rate of $3 \times 10^{-4}$ for 500,000 steps with batch size 256. During training images are cropped randomly to size of 512$\times$512. For evaluation, the random crop is replaced by a center crop but in the end metrics are calculated on a further center crop of 384$\times$384. All the models are initialized with the checkpoints of a model trained on ImageNet \cite{imagenet}.

\subsection{Handling of multiple timeframes}

We explored two options for handling multiple Sentinel-2 timeframes in the model. In the first one we concatenated all timeframes in the channel dimension. In the second approach we passed each timeframe through a shared encoder (see \Cref{sec:model}). The second approach seems to require cross-time information fusion (see \Cref{sec:cross_time_fusion}) and/or a pairing scheme for timeframes (see \Cref{sec:pairing_schemes}) to exceed the performance of the first approach (see \Cref{tab:timeframe_handling_experiments}). Both approaches used slightly modified off-the-shelf HRNet \cite{hrnet} and U-Net \cite{unet}. See \Cref{sec:model_encoder} for a description of how HRNet was modified, in case of U-Net the modifications were equivalent. Unless stated otherwise, in all remaining experiments we use the second approach with cross-time information fusion and no pairing scheme.

\begin{table}
    \caption{\label{tab:timeframe_handling_experiments}Comparison of two approaches to handling multiple timeframes: concatenating the timeframes in the channel dimension and separate encoding using a shared encoder. The second approach requires additional modifications to exceed the performance of the first approach.}
    \centering
        \begin{tabular}{lr}
        \\
        \toprule
            Model    & mIoU (Buildings) \\
        \midrule
            U-Net concatenate timeframes in channel dimension & 72.7  \\
            HRNet concatenate timeframes in channel dimension & 73.8 \\
            HRNet separate timeframe encoding & 70.7 \\
            HRNet separate timeframe encoding + cross-time fusion  & 76.7 \\
            HRNet separate timeframe encoding + Pairing (17th frame) & \textbf{76.9} \\
            HRNet separate timeframe encoding + Cross-time fusion + Pairing (17th frame) & 76.4 \\
        \bottomrule
        \end{tabular}
\end{table}

\subsection{Pairing schemes} \label{sec:pairing_schemes}

Inspired by HighResNet \cite{deudon2020highres}, where each low-resolution frame is paired with the per-pixel median of the stack, we experimented with various pairing schemes.
We find that pairing each timeframe with the 17th timeframe (that is closest in time to the teacher label) provides a significant boost in performance over model without pairing. 
In comparison, model trained with no pairing but with cross-time communication (\Cref{sec:cross_time_fusion}) performs slightly worse (see \Cref{tab:timeframe_handling_experiments}).
Surprisingly enough, model trained with both pairing (17th frame) and cross-time communication performs worse than that with just pairing.
We also experimented with pairing with either the median (as is done in HighResNet), mean or second darkest timeframe and find none of these pairing schemes to be better than pairing with the 17th frame.
Ultimately, we used cross-time information fusion and no pairing scheme, based on the observation that this method appeared more robust in cases where the 17th frame was cloudy.

\subsection{Resolution sensitivity analysis}

To understand the influence of input, output and target resolutions on model performance we carry out the following resolution sensitivity experiments:

\paragraph{Input resolution}

We altered the resolution of the input while keeping the model output and label at 50 cm resolutions. The effective resolution of the Sentinel-2 input is 10 m or lower depending on the specific band, but we find that upsampling it with a simple image resize has non-trivial impact on model performance. For example, training on 4 meter inputs provides a sizable improvement over training on 8 meter inputs (\Cref{tab:sentinel-input-resolution}). However, increasing the input resolution to 2 meters does not provide similar benefit. The models in \Cref{tab:sentinel-input-resolution} are trained on eight Sentinel-2 timeframes because of the computational constraints associated with training at 2 meter input resolution. For training on the stack of 32 images, we only explored training on 8 and 4 meter inputs. The performance improvement associated with training on 4 meter inputs is also observed in this case.

\begin{table}
    \caption{\label{tab:sentinel-input-resolution}Sensitivity analysis of the Sentinel-2 input resolution (8 timeframes).}
    \centering
        \begin{tabular}{lr}
        \\
        \toprule
            Input resolution    & mIoU (Buildings) \\
        \midrule
            8 meters & 74.3 \\
            4 meters & 75.6 \\
            2 meters & 75.8 \\
        \bottomrule
        \end{tabular}
\end{table}

\paragraph{Output resolution}

For output resolution sensitivity experiments we alter the effective resolution of the output while keeping the model input at 4 m and the label at 50 cm resolutions.

As described in \Cref{sec:model}, to produce super-resolved outputs we add residual upsampling blocks akin to the decoder blocks used in U-Net to progressively upsample the input. We investigate the usefulness of these blocks by progressively replacing them with a bi-linear resize operation. Although the output of the model is at the same resolution as the label, the effective resolution is actually much lower since the resize operation does not add any new details. \Cref{tab:output-resolution-sensitivity} shows that using the upsampling block to increase the effective resolution leads to an increase in the performance of the model. At 4 meters, no upsampling blocks are used, resulting in lower performance compared to other models that make use of them.

\begin{table}
    \caption{\label{tab:output-resolution-sensitivity}Sensitivity analysis of the Sentinel-2 output resolution demonstrating usefulness of upsampling blocks (32 timeframes).}
    \centering
        \begin{tabular}{lr}
        \\
        \toprule
            Effective output resolution    & mIoU (Buildings) \\
        \midrule
            4 meters  & 76.1  \\
            2 meters & 76.6 \\
            1 meter & 76.8 \\
            50 centimeters & 77.3 \\
        \bottomrule
        \end{tabular}
\end{table}

\paragraph{Label resolution}

For label resolution sensitivity experiment we alter the effective resolution of the label while keeping the model input at 4 m and output at 50 cm resolutions. Unsurprisingly, we see that the model performs better when trained on higher resolution labels, as shown in \Cref{tab:under-resolve-labels}.

\begin{table}
    \caption{\label{tab:under-resolve-labels}Sensitivity analysis of label resolution (32 timeframes).}
    \centering
        \begin{tabular}{lr}
        \\
        \toprule
            Effective label resolution  & mIoU (Buildings) \\
        \midrule
            4 meters & 76.2 \\
            2 meters & 75.6 \\
            50 centimeters & 76.8 \\
        \bottomrule
        \end{tabular}
\end{table}

\paragraph{Comparison to single-frame detection with varying image resolution}

We wished to understand the performance of our model in comparison to a high-resolution building detection model operating on a single frame of imagery. That is, if only one image was available, what resolution would it need to be in order to get the same accuracy of detection as we obtain with 32 Sentinel-2 frames.

To do this, we trained a model similar to our teacher model on downsampled high-resolution images. The images are downsampled to a target lower resolution and resized back to the original image dimensions. Thus the dimensions of the images are the same but the information content is reduced. We find that at 4 meter resolution, this model's performance matches the performance of the Sentinel-2 super-resolution model (\Cref{tab:high-resolution-sensitivity}). The 50 cm model has comparable metrics to the model that was used to generate teacher labels for training the Sentinel-2 models.

\begin{table}
    \caption{\label{tab:high-resolution-sensitivity}Resolution sensitivity for a model trained on a single high-resolution image. For comparison, the last row represents our best Sentinel-2 super-resolution segmentation model. It has comparable mIoU to the 4 meter model.}
    \centering
        \begin{tabular}{lr}
        \\
        \toprule
            Effective image resolution (cm)  & mIoU (Buildings) \\
        \midrule
            1000 & 67.4 \\
            500 & 75.8 \\
            \textbf{400} & \textbf{78.1} \\
            300 & 80.5 \\
            200 & 82.9 \\
            100 & 84.8 \\
            50 & 85.5 \\
        \midrule
            Best Sentinel-2 model & \textbf{79.0} \\
        \bottomrule
        \end{tabular}
\end{table}

\subsection{Training set size}

To quantify the impact of dataset size on model performance we trained our model on subsets of training data of different sizes. \Cref{tab:training-set-size} shows monotonic improvement in performance as training data size increases. The model trained on the full set of 10 million images outperforms the model trained on 10\% of the training set by 2 percentage points.

\begin{table}
    \caption{\label{tab:training-set-size}Building segmentation performance as a function of fraction of training data size used (32 timeframes).}
    \centering
        \begin{tabular}{lr}
        \\
        \toprule
            Fraction of training data  & mIoU (Buildings) \\
        \midrule
            1\% & 69.1 \\
            5\% & 73.4 \\
            10\% & 74.7 \\
            100\% & \textbf{76.6} \\
        \bottomrule
        \end{tabular}
\end{table}

\subsection{Number and position of Sentinel-2 timeframes}

Each training example contains 32 Sentinel-2 timeframes with the corresponding high-resolution image located temporally somewhere between the 16th and 17th timeframes. The timeframes are sorted by time. This means that the model gets to see 16 frames in the past and 16 in the future around the high-resolution image time. We carry out a series of experiments to determine the relationship between the number of timeframes and the performance of the model.

We find that increasing the number of timeframes leads to a monotonic increase in the performance of the model. In \Cref{tab:number-of-timeframes}, the model trained on all 32 timeframes outperforms the model trained on a single timeframe by 5 percentage points. As a corollary to this, we train a model on a single timeframe duplicated 32 timeframes. This model achieves a mean intersection over union of 71.7\% compared to 76.7\% of the model trained on all timeframes. This shows that the additional timeframes provide useful information to the model. The conditions at the time a snapshot is taken is different for the 32 timeframes, thus there must be features that are captured in some frames but absent in others. Using multiple frames allows the model to draw from the information available in each time frame to make a single prediction. The use of multiple frames can also be understood as producing a `dither' effect. Since noise should be randomly distributed across the image, shifts between consecutive frames allow the model to isolate the signal from the noise to produce a super-resolved output.

\begin{table}
    \caption{\label{tab:number-of-timeframes}Model performance as a function of the number of timeframes in the Sentinel-2 stack.}
    \centering
        \begin{tabular}{lr}
        \\
        \toprule
            Number of timeframes    & mIoU (Buildings) \\
        \midrule
            1  & 71.7 \\
            2  & 73.2 \\
            4  & 74.5 \\
            8  & 75.8 \\
            16  & 76.1 \\
            32  & \textbf{76.7} \\
        \bottomrule
        \end{tabular}
\end{table}

We also explore how sensitive the model is to having access to future or past timeframes by training the model on only past or future frames. In \Cref{tab:temporal-relationship-of-timeframes}, the model that only gets to see 16 past timeframes is better than a model that sees only 16 future timeframes, and roughly comparable to one that sees 8 past and 8 future timeframes. 

\begin{table}
    \caption{\label{tab:temporal-relationship-of-timeframes}Effect of the temporal relationship between the high-resolution label and the Sentinel-2 stack, with 16 frames either at the end of the stack (after the time of the high-resolution label), the start (before the label), or split across both.}
    \centering
        \begin{tabular}{lr}
        \\
        \toprule
            Method    & mIoU (Buildings) \\
        \midrule
            Only future  & 75.8  \\
            Only past  & \textbf{76.6}
            \\
            Equal past \& future & 76.3 \\
        \bottomrule
        \end{tabular}
\end{table}

\subsection{Building counts via centroid prediction}
\label{sec:experiments/extension_to_centroids}

\begin{figure}
    \centering
    \includegraphics[width=.85\textwidth]{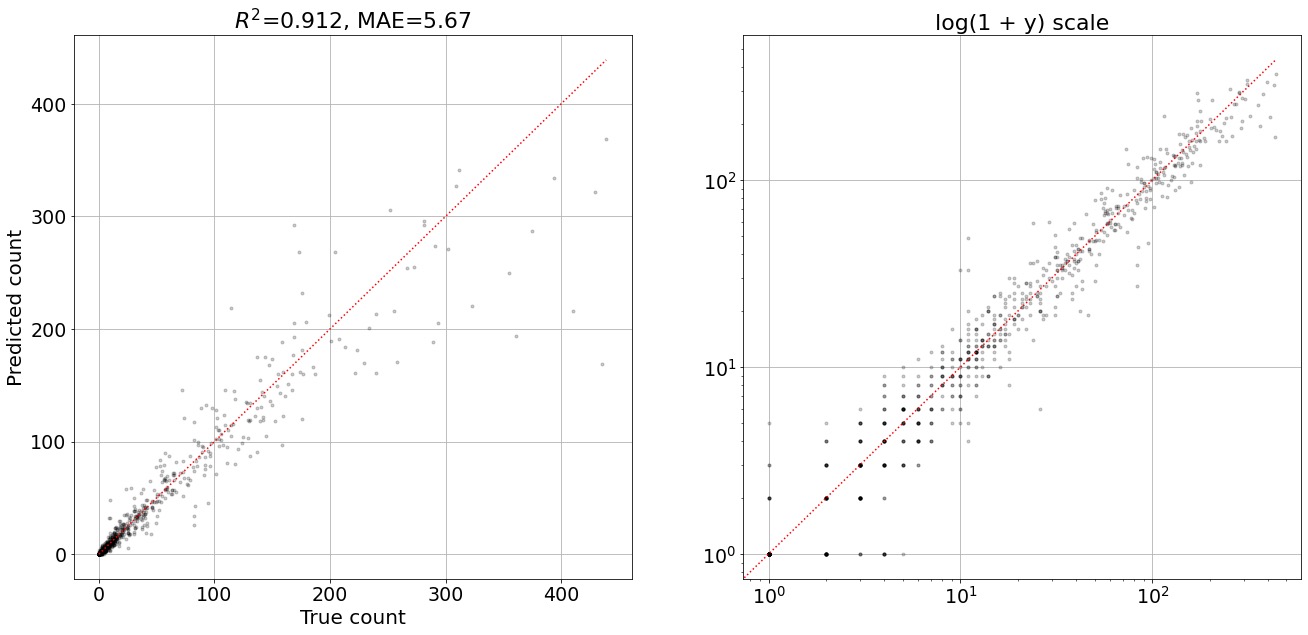}
    \caption{\label{fig:counts_comparison}Regression plots (in linear and logarithmic scale) and metrics for the buildings count from just a stack of low-resolution images. Counts are made on tiles of size $192^2$ m$^2$ on the ground.}
\end{figure}

On the auxiliary task of tile level count prediction via centroid prediction as outlined in \Cref{sec:hi_res_teacher_model/extension_to_centroids} we observe a very high correlation between predicted building count and the true count (see \Cref{fig:counts_comparison}). We were able to achieve a coefficient of determination ($R^2$) of 0.912 and MAE of 5.67 on human labels. For comparison, model trained to do the same but using 50 cm imagery was able to achieve $R^2$ of 0.955 and MAE of 4.42 on human labels.

\subsection{Incidence angle metadata for high-resolution imagery}
\label{sec:experiments/high_res_incidence_angle}

\begin{figure}
\centering
\begin{minipage}{.45\textwidth}
  \centering
  \includegraphics[height=.15\textheight]{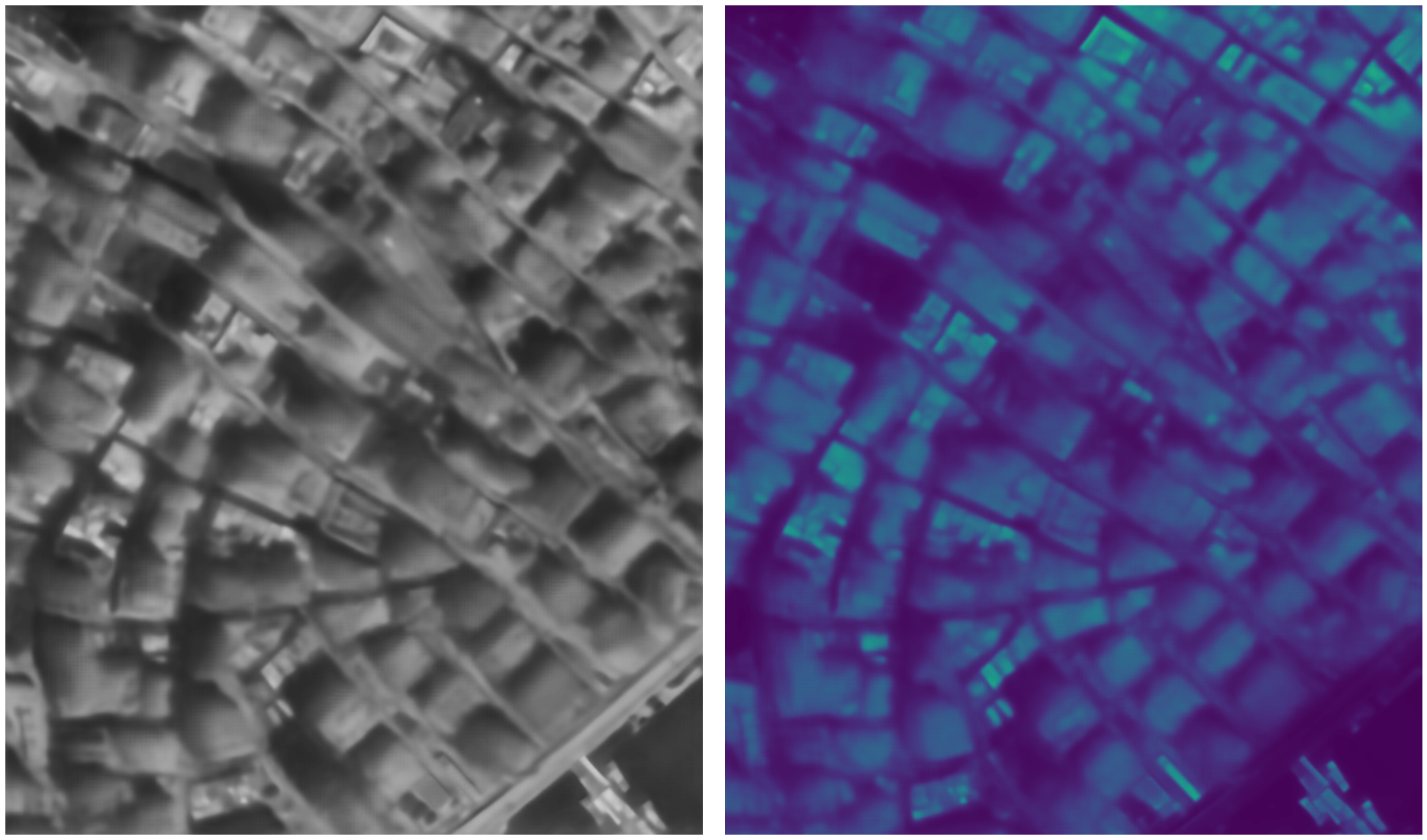}
  \captionof{figure}{\label{fig:without_highres_incidence_angle}Output from a model trained \textbf{without} high-resolution incidence metadata in Manhattan.}
\end{minipage}
\qquad
\begin{minipage}{.45\textwidth}
  \centering
  \includegraphics[height=.15\textheight]{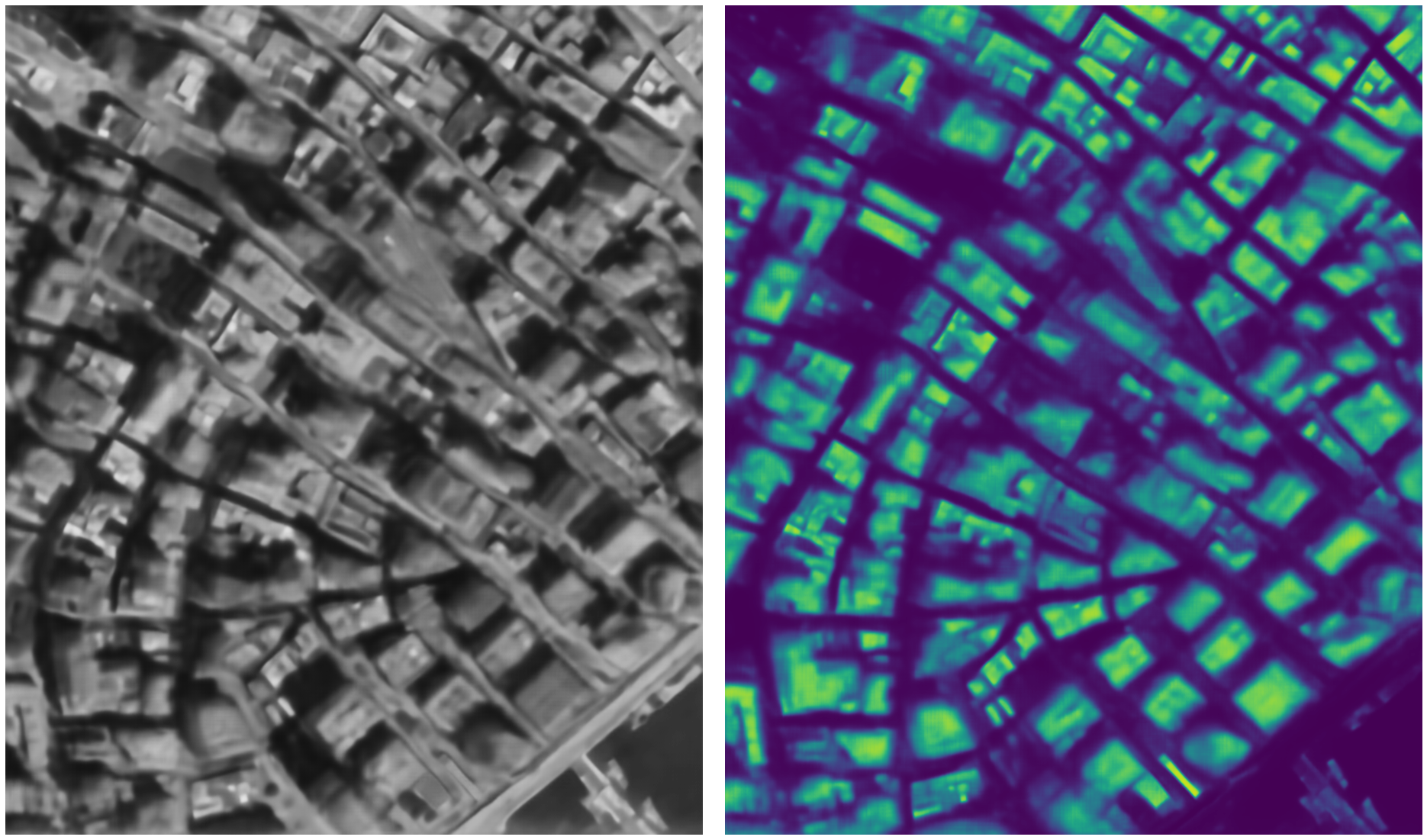}
  \captionof{figure}{\label{fig:with_highres_incidence_angle}Output from a model trained \textbf{with} high-resolution incidence metadata in Manhattan.}
\end{minipage}
\end{figure}

As mentioned in \Cref{sec:orthorectification}, 80\% of the high-resolution images we use are not orthorectified. This is a problem especially for tall buildings as the student model is not able to guess the position of the roofs in the high-resolution image from which the teacher label was generated and hence tends to produce blurry output in those cases. To mitigate this issue during training we pass the high-resolution satellite image incidence angle metadata on input to the student model, so that it can learn to modify its output to match the teacher label. During inference we set the incidence angle to $0^\circ$ so that the model predicts orthorectified output. See \Cref{fig:without_highres_incidence_angle} and \Cref{fig:with_highres_incidence_angle} for an example of visual quality improvement in model output.

\subsection{Building height prediction}
\label{sec:experiments/extension_to_building_heights}

As described in \Cref{sec:height_labels}, 20\% of examples in the datasets have per pixel above-ground object height labels that we used to train the model to solve the auxiliary task of building height prediction (capped at 100 meters). The best model was able to achieve mean absolute error of 1.5 meters on the building instance height validation dataset. Additional evaluation results for height prediction can be seen in \Cref{tab:height-prediction}. Further research is necessary to understand what cues the model uses for height prediction; for example this may be related to shadow length, or a parallax effect due to Sentinel-2 channels being captured at slightly different times as the satellite passes overhead.

\begin{table}
    \caption{\label{tab:height-prediction}Mean and percentiles of building instance height absolute error (AE) of the best model on the height prediction validation dataset.}
    \centering
        \begin{tabular}{lrrrrr}
        \\
        \toprule
            Building height range & Mean AE & 50th pct. AE & 90th pct. AE & 95th pct. AE & 99th pct. AE \\
        \midrule
            (0 m; 5 m) & 1.01 m & 0.76 m & 2.15 m & 2.71 m & 4.54 m \\
            {[}5 m; 20 m) & 1.63 m & 1.14 m & 3.71 m & 4.86 m & 7.87 m \\
            {[}20 m; 100 m{]} & 3.64 m & 2.53 m & 7.95 m & 11.00 m & 17.95 m \\
            (0 m; 100 m{]} & 1.50 m & 0.99 m & 3.35 m & 4.66 m & 8.48 m \\
        \bottomrule
        \end{tabular}
\end{table}

\subsection{Performance across different validation datasets}

We evaluated the best model across all validation datasets on building segmentation using mean Intersection over Union (mIoU) metric, on building counting using the coefficient of determination ($R^2$) metric and on total built-up area prediction using relative error metric. For comparison, we included mIoU of the teacher model on the high-resolution images. Results are summarized in \Cref{tab:metrics-across-regions} and the score thresholds used for binarization that maximize the metrics are in \Cref{tab:metrics-across-regions-score-thresholds}. The model performs well across all validation datasets with notable exception of building counting on the "Africa hard" dataset. The gap in performance between the Sentinel-2 model and the high-resolution model varies 3.3-6.5 mIoU and is the largest on the Africa validation dataset. Moreover, the score thresholds are similar across all validation datasets.

\begin{table}
    \caption{\label{tab:metrics-across-regions}The best model evaluated on all building segmentation validation datasets.}
    \centering
        \begin{tabular}{lrrrrr}
        \\
        \toprule
            Dataset & Teacher model mIoU & mIoU & Building count $R^2$ & Built-up area error \\
        \midrule
            Africa & 85.5 & 79.0 & 0.912 & 0.79\% \\
            Africa hard & 85.2 & 80.0 & 0.837 & 0.15\% \\
            South-Southeast Asia & 87.2 & 82.7 & 0.936 & 0.79\% \\
            Latin America & 87.7 & 84.4  & 0.887 & 0.18\%\\
        \bottomrule
        \end{tabular}
\end{table}

\begin{table}
    \caption{\label{tab:metrics-across-regions-score-thresholds}Score thresholds used for binarization that maximize the metrics for the best model.}
    \centering
        \begin{tabular}{lrrrrr}
        \\
        \toprule
            Dataset & Best mIoU score threshold & Best built-up area error score threshold \\
        \midrule
            Africa & 0.34 & 0.40 \\
            Africa hard & 0.37 & 0.41 \\
            South-Southeast Asia & 0.34 & 0.41 \\
            Latin America & 0.37 & 0.45 \\
        \bottomrule
        \end{tabular}
\end{table}

\section{Discussion} 

In this work, we present an end-to-end super-resolution segmentation framework to segment buildings and roads from Sentinel-2 imagery at a much higher effective resolution than the input imagery. To this end we demonstrate label transfer from a high-resolution satellite to a low-resolution satellite that both image the same surface of the earth. We show that such label transfer not only allows generation of more accurate and fine detailed labels but also enables automatic label generation through a teacher model trained to perform the same task but using imagery from a high-resolution satellite. This significantly reduces the amount of high-resolution imagery needed for certain analysis tasks, such as large scale building mapping.

Unfortunately, our approach has some limitations. First, it still assumes that one has access to a good amount of high-resolution images to first train a teacher high-resolution model and then also to generate a large dataset to train a student model as described in \Cref{sec:dataset}. However, this is a one-time cost. Additionally, with the advancement in deep learning, models can now be trained in a much more data efficient manner \cite{cong2022satmae}. Furthermore, there are many publicly available datasets and models that can be used to bootstrap different tasks \cite{spacenetrepo}. Second, it assumes that for a given location one can assemble a deep stack (i.e. 32) of cloud-free Sentinel-2 images. This actually can be quite problematic if the stack has to be centered timewise around a fairly old or recent date, and especially if the location is cloudy; in humid locations such as Equatorial Guinea we noticed that for some very cloudy locations a 32 timeframe stack can span more than 2 years.

The results we present in this technical report are preliminary and we believe that there is still room for improvement. For example, we did not evaluate the model on change detection, and do not have metrics on how many Sentinel-2 timeframes with change does the model need to see to detect these changes consistently. Additionally, recent advances in super-resolution using generative AI models such as GANs \cite{pineda2020generative} and diffusion models \cite{saharia2021image} have shown potential. These approaches have however been limited to single image super resolution of natural images and have not been extensively used for remote sensing tasks where multiple low-resolution images of the same location are available. As such, a promising area of future research could be exploration of use of some of these approaches for multi-frame super-resolution segmentation tasks in remote sensing.

\subsection{Social and ethical considerations}

We believe that timely and accurate building information, particularly in areas which have few mapping resources already, are critically important for disaster response, service delivery planning and many other beneficial applications. However, there are potential issues with improvements in remote sensing analysis, both in terms of unintended consequences and from malicious use. Where such a model is used as a source about information on human population centres, for example during emergency response in a poorly mapped and inaccessible area, any false negatives could lead to settlements being neglected, and false positives lead to resources being wasted. Particular risks for the kind of Sentinel-2 based analysis we describe include settlements consisting of very small buildings made of natural materials, and buildings in deserts, both of which are challenging.

\section{Acknowledgements}

We thank Sergii Kashubin, Maxim Neumann and Daniel Keysers for feedback which helped to improve this paper.

\bibliographystyle{unsrt}
\bibliography{main}

\section{Appendix}
\label{sec:appendix}

\begin{figure}[!htb]
  \centering
  \includegraphics[width=\textwidth]{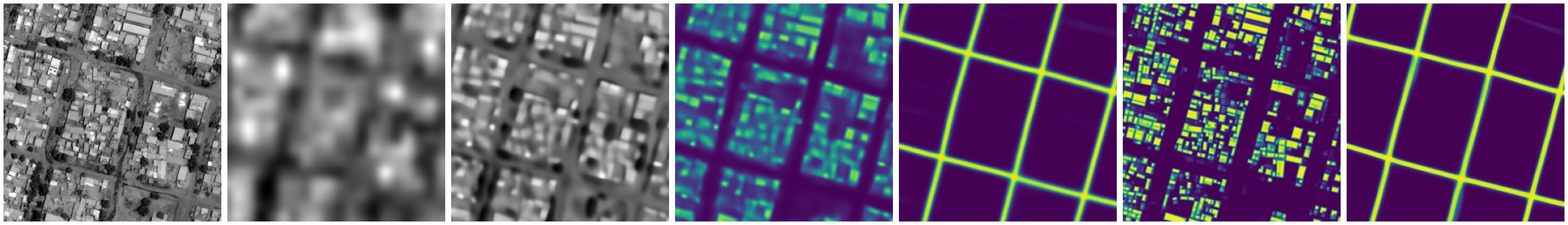}
  \includegraphics[width=\textwidth]{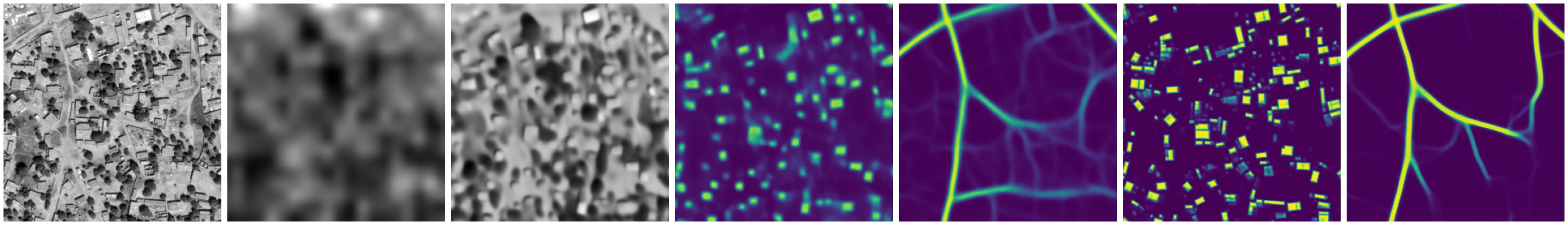}
  \includegraphics[width=\textwidth]{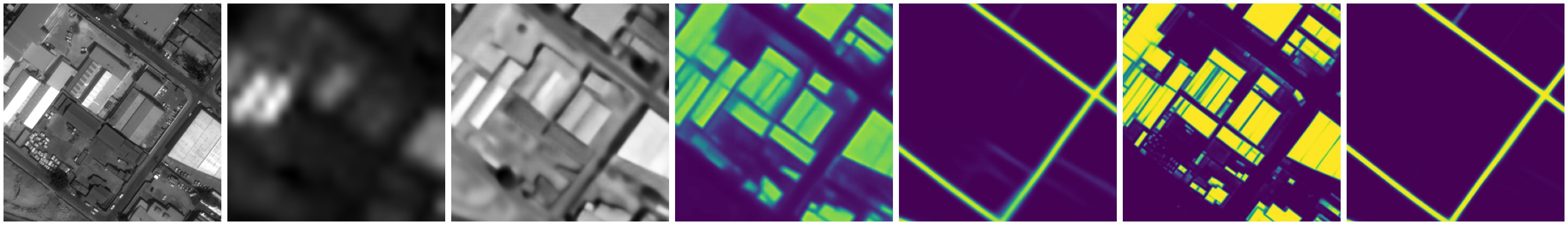}
  \includegraphics[width=\textwidth]{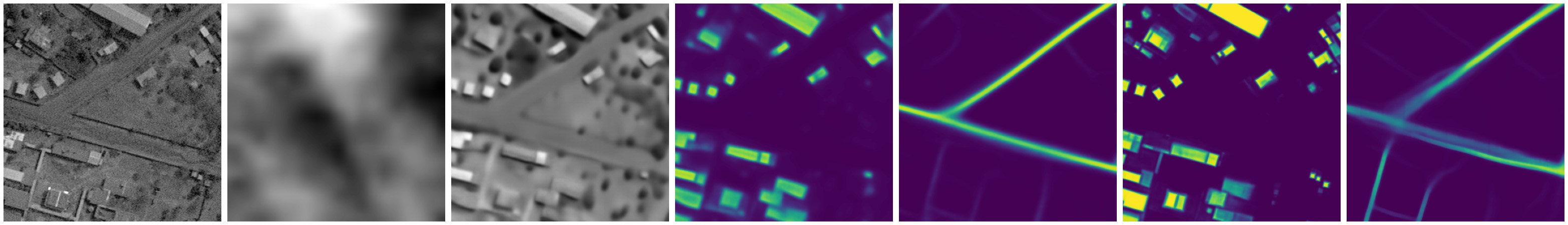}
  \includegraphics[width=\textwidth]{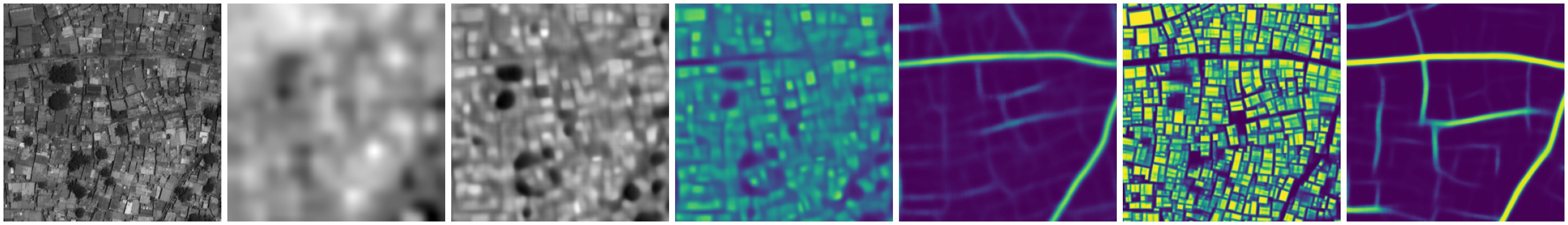}
  \includegraphics[width=\textwidth]{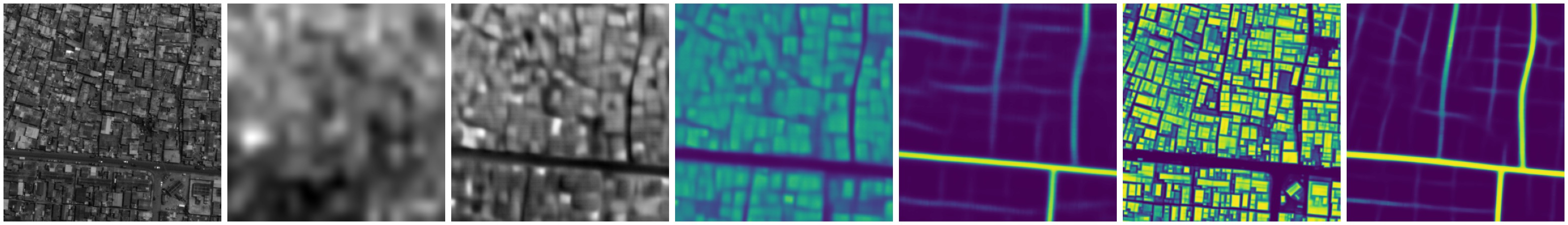}
  \includegraphics[width=\textwidth]{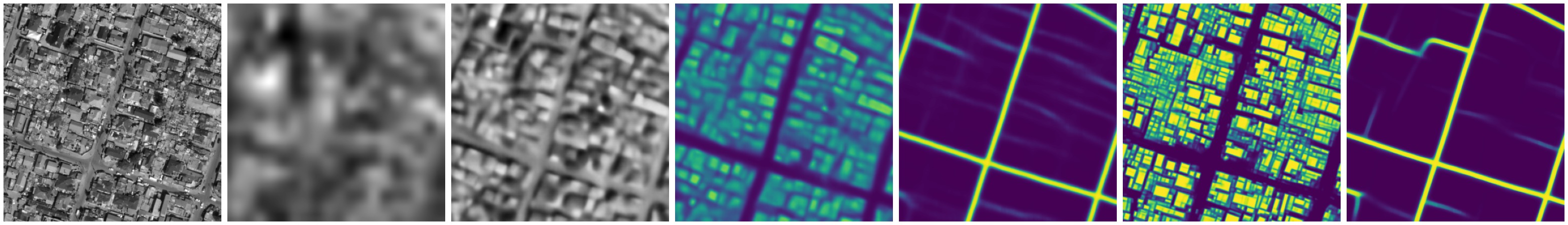}
  \includegraphics[width=\textwidth]{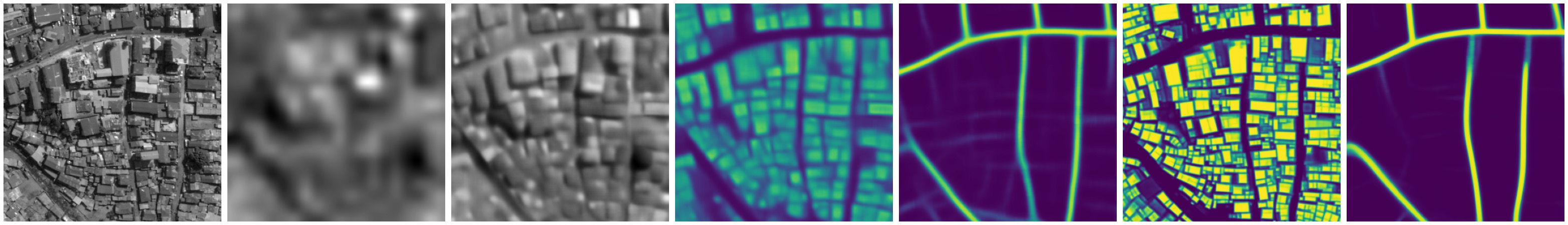}
    
  \begin{tabular}{*{7}{C{0.117\textwidth}}}
     \footnotesize 50 cm grayscale & \footnotesize Sentinel-2 grayscale & \footnotesize Sentinel-2 grayscale super-resolution & \footnotesize Building detection & \footnotesize Road detection & \footnotesize Teacher building detection & \footnotesize Teacher road detection \\
  \end{tabular}

  \caption{Urban examples of building and road detection output, each covering an area of $192^2$ m$^2$.}

\end{figure}

\begin{figure}
  \centering
  \includegraphics[width=\textwidth]{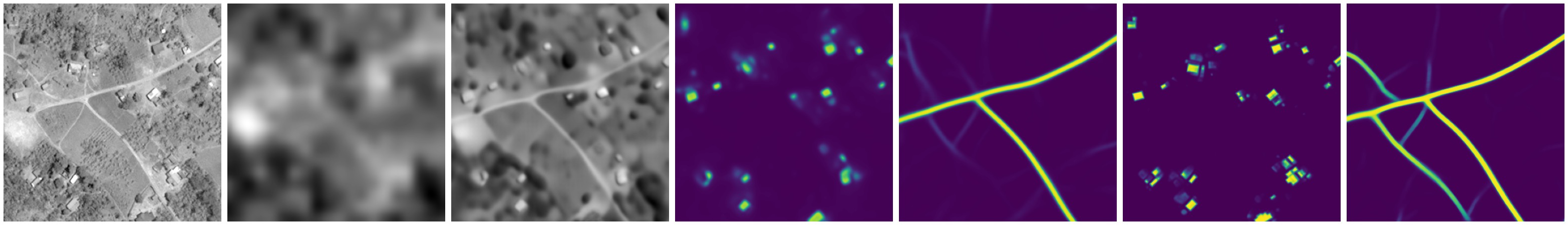}
  \includegraphics[width=\textwidth]{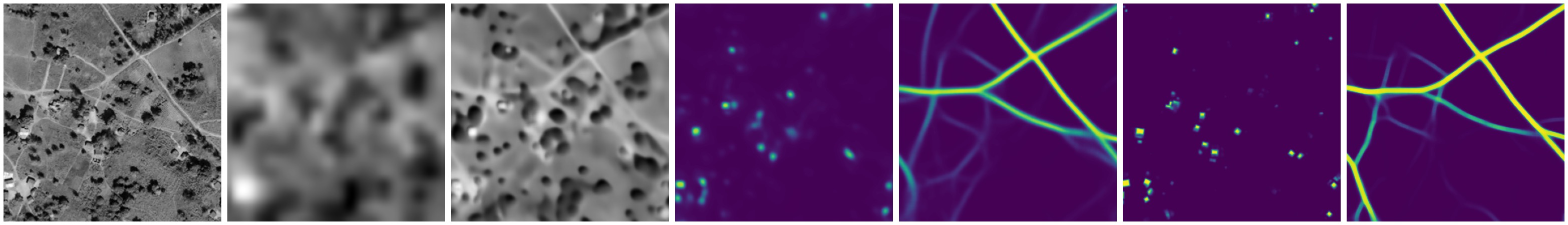}
  \includegraphics[width=\textwidth]{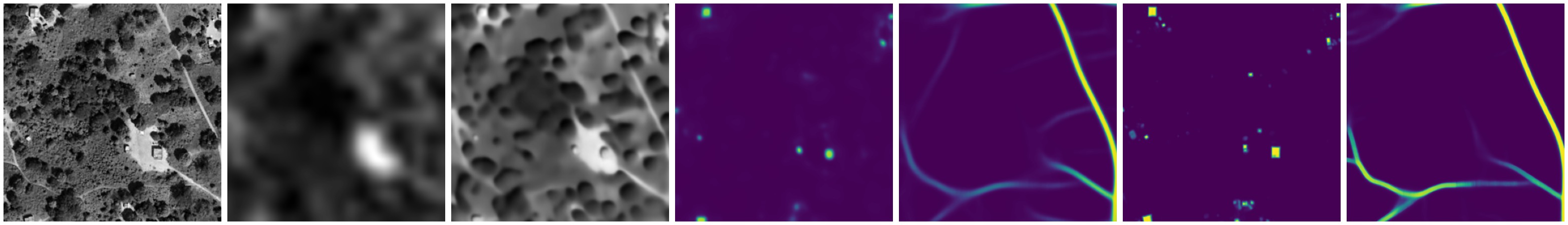}
  \includegraphics[width=\textwidth]{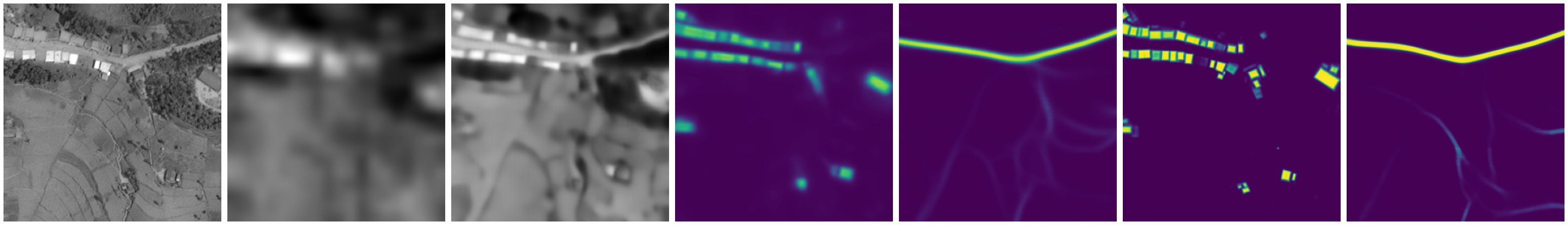}
  \includegraphics[width=\textwidth]{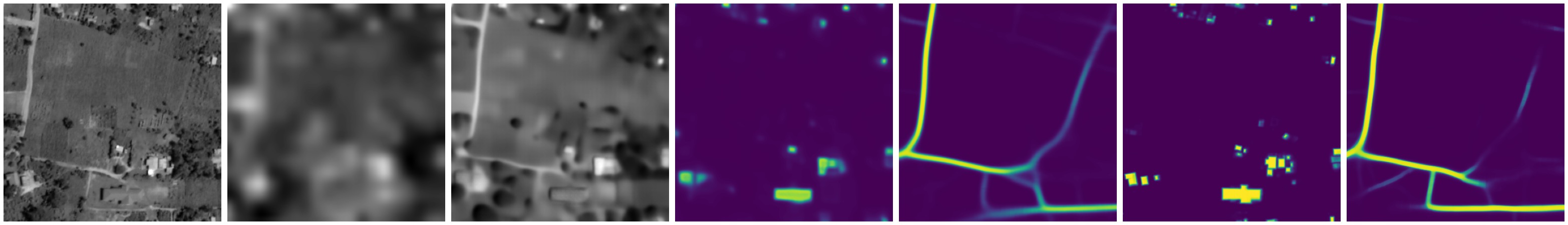}
  \includegraphics[width=\textwidth]{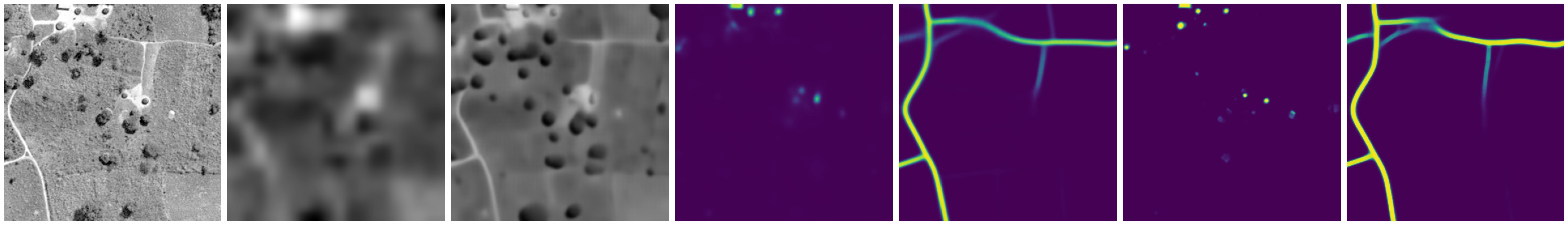}
  \includegraphics[width=\textwidth]{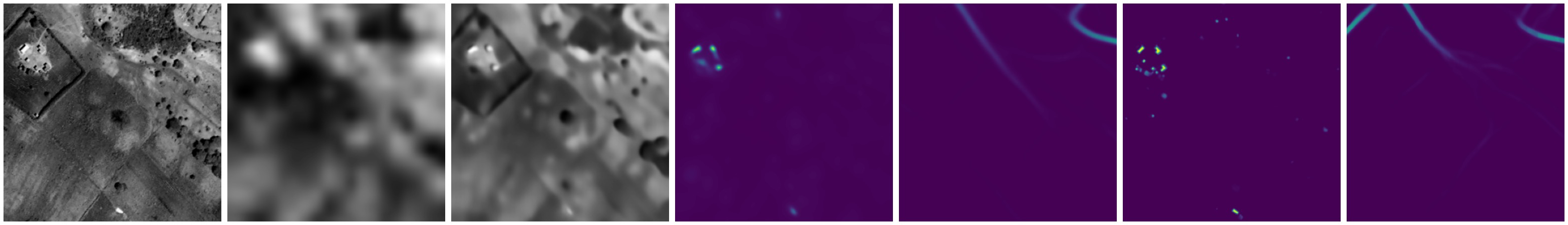}
  \includegraphics[width=\textwidth]{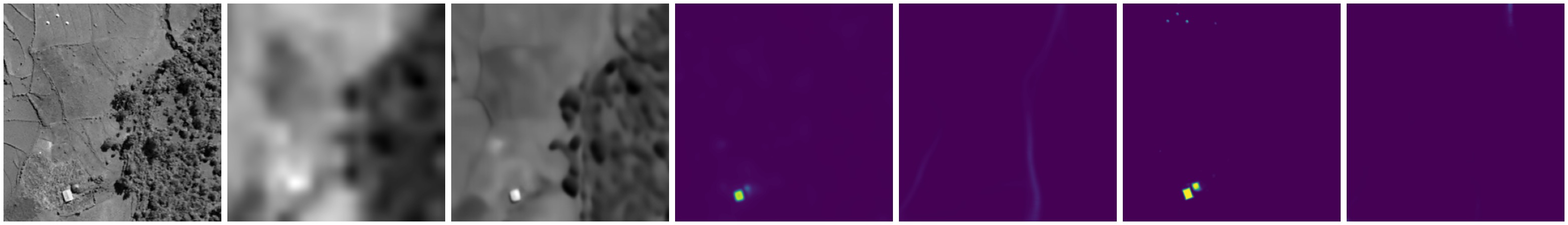}
    
  \begin{tabular}{*{7}{C{0.117\textwidth}}}
     \footnotesize 50 cm grayscale & \footnotesize Sentinel-2 grayscale & \footnotesize Sentinel-2 grayscale super-resolution & \footnotesize Building detection & \footnotesize Road detection & \footnotesize Teacher building detection & \footnotesize Teacher road detection \\
  \end{tabular}

  \caption{Rural examples of building and road detection output, each covering an area of $192^2$ m$^2$.}

\end{figure}

\begin{figure}
  \centering
  \begin{tabular}{m{0.03\textwidth}m{0.9\textwidth}}
  (a) & \includegraphics[width=0.9\textwidth]{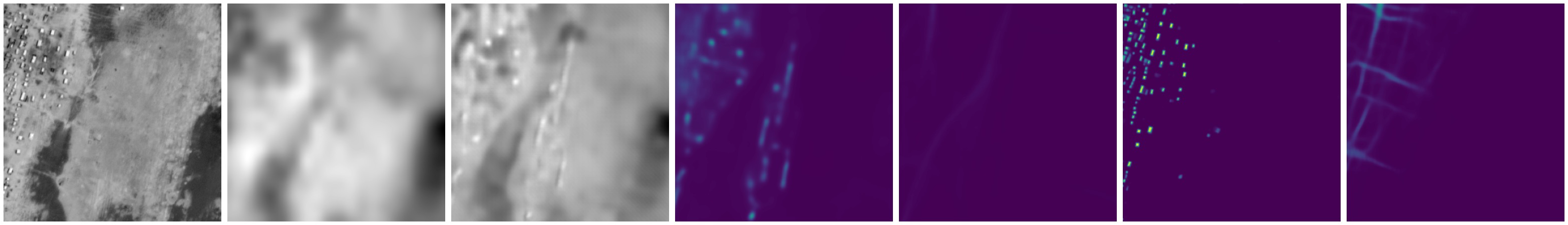} \\
  (b) & \includegraphics[width=0.9\textwidth]{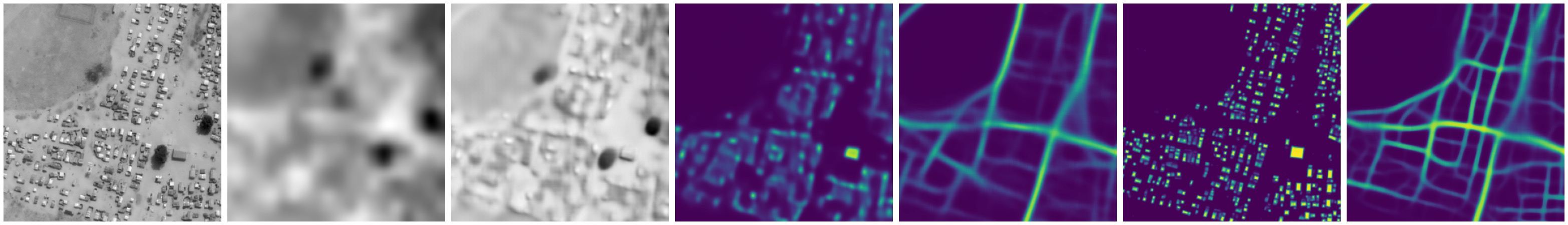} \\
  (c) & \includegraphics[width=0.9\textwidth]{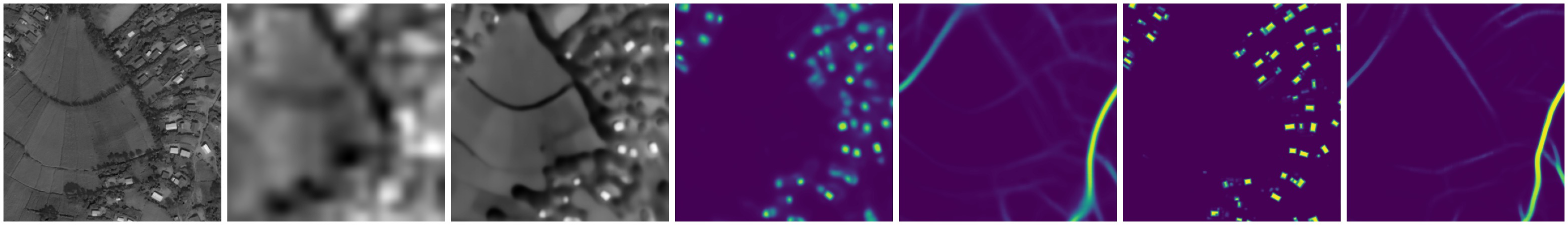} \\
  (d) & \includegraphics[width=0.9\textwidth]{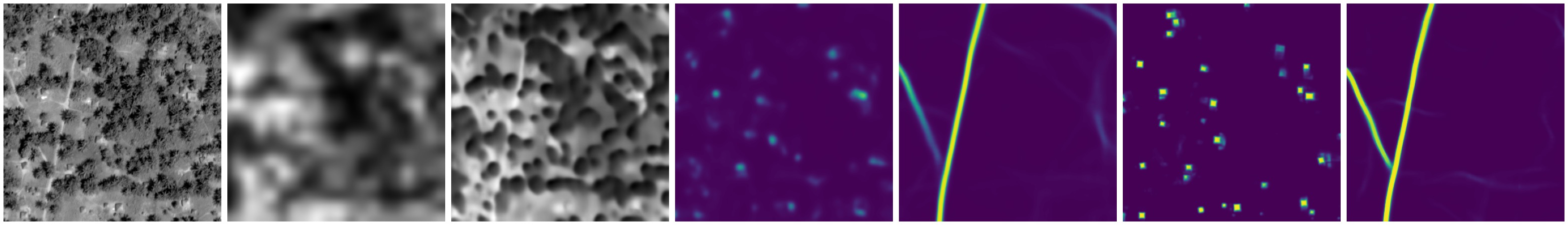} \\
  (e) & \includegraphics[width=0.9\textwidth]{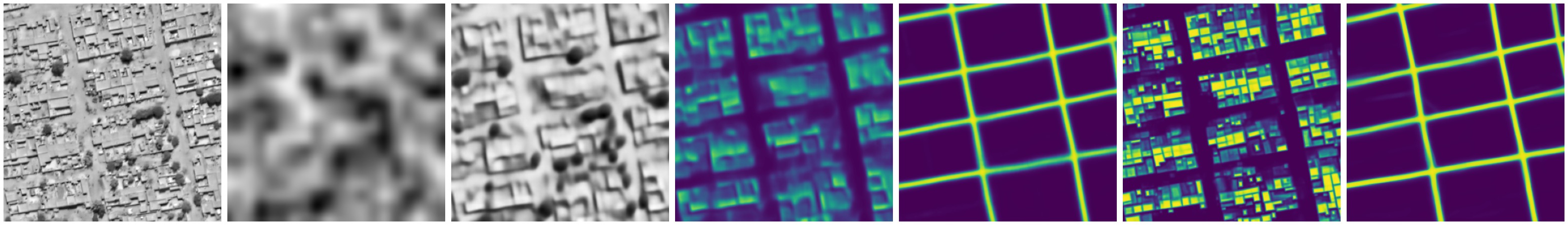} \\
  (f) & \includegraphics[width=0.9\textwidth]{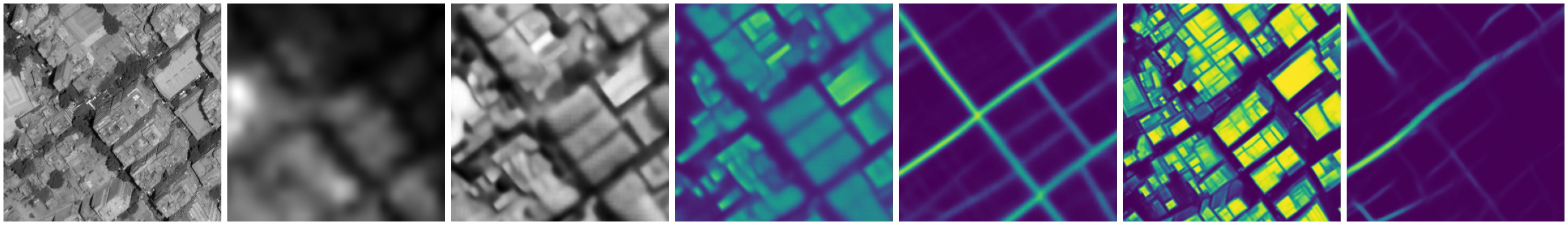} \\
  (g) & \includegraphics[width=0.9\textwidth]{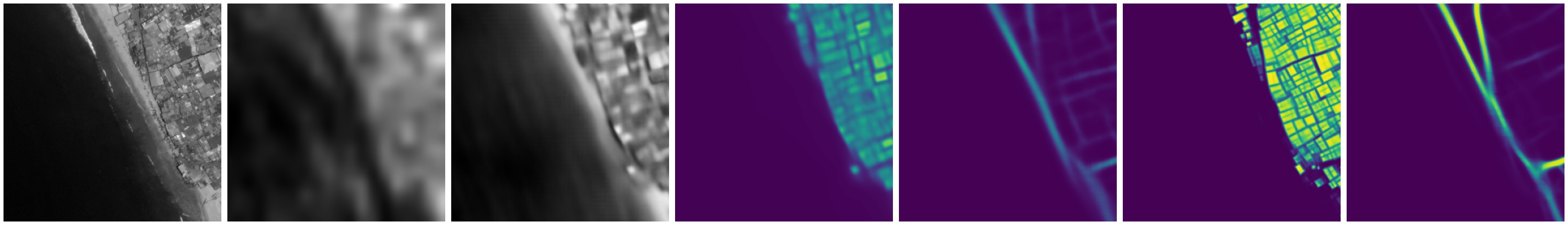} \\
  \end{tabular}
    
  \begin{tabular}{C{0.03\textwidth}*{7}{C{0.103\textwidth}}}
     & \footnotesize 50 cm grayscale & \footnotesize Sentinel-2 grayscale & \footnotesize Sentinel-2 grayscale super-resolution & \footnotesize Building detection & \footnotesize Road detection & \footnotesize Teacher building detection & \footnotesize Teacher road detection \\
  \end{tabular}

  \caption{\label{fig:interesting-examples}Interesting examples of building and road detection output, each covering an area of $192^2$ m$^2$: (a) some regions can develop very rapidly, the 50 cm image is 3 weeks older than the next available cloud-free Sentinel-2 image but new structures have already appeared during that timespan; (b) the model struggles with separating many small building structures from each other; (c) the model can be sensitive to roof color, in this example lighter colored roofs have higher confidence; (d) model struggles on small buildings interspersed with trees; (e) buildings in desert environment can be challenging; (f) tall structures can have blurry predictions due to 50 cm imagery often being at high off-nadir angle; (g) the road prediction can sometimes be better than the teacher road prediction, in this case the model is less confident the beach is a road.}
\end{figure}

\begin{figure}
  \centering
  \begin{tabular}{m{0.07\textwidth}m{0.3\textwidth}m{0.3\textwidth}m{0.3\textwidth}}
  \scriptsize 2017-06-02 & \includegraphics[width=0.3\textwidth]{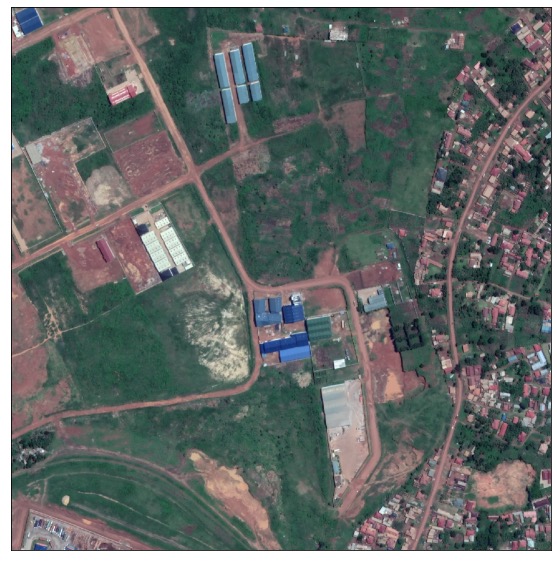} & \includegraphics[width=0.3\textwidth]{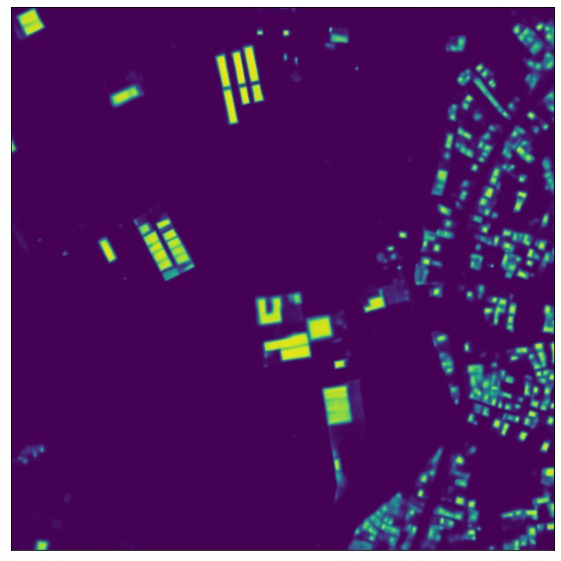} & \includegraphics[width=0.3\textwidth]{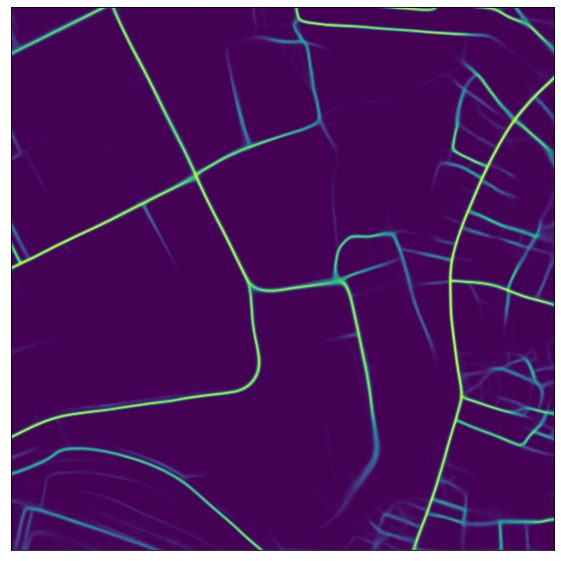} \\
  \scriptsize 2019-12-19 & \includegraphics[width=0.3\textwidth]{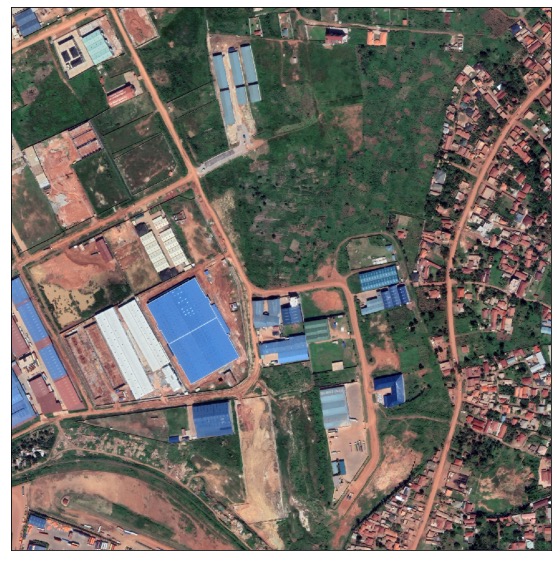} & \includegraphics[width=0.3\textwidth]{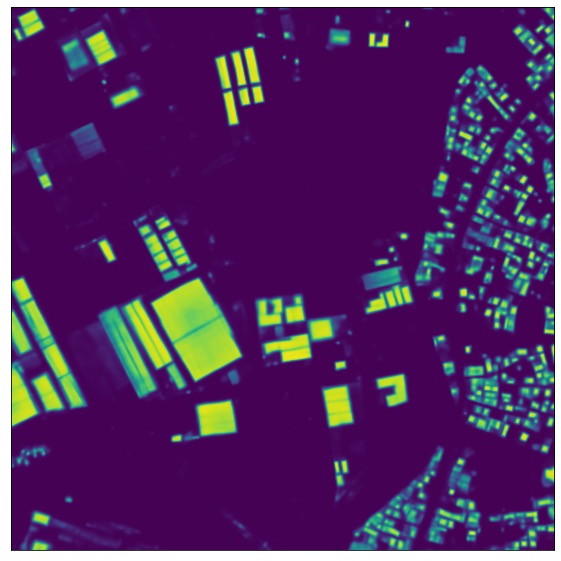} & \includegraphics[width=0.3\textwidth]{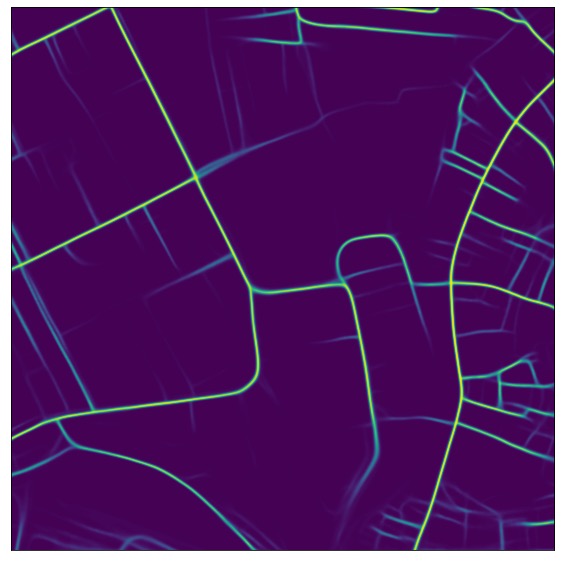} \\
  \scriptsize 2021-04-03 & \includegraphics[width=0.3\textwidth]{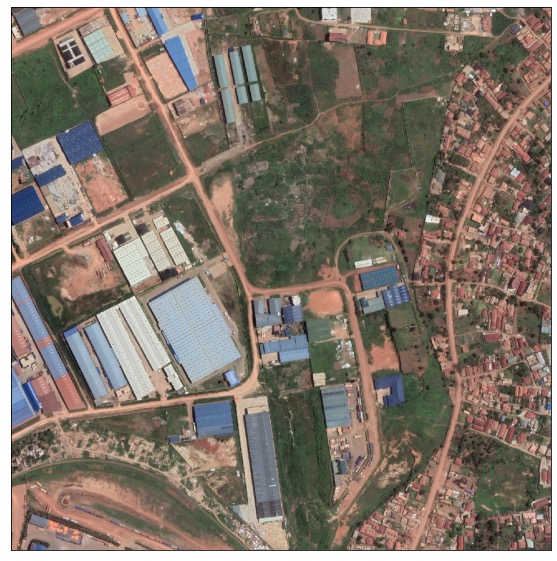} & \includegraphics[width=0.3\textwidth]{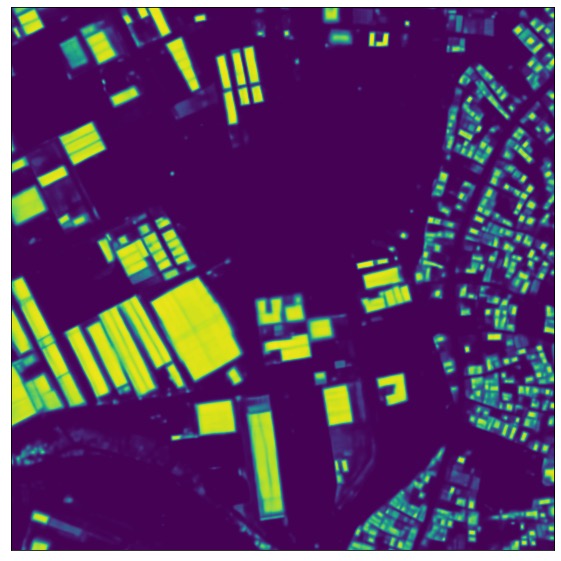} & \includegraphics[width=0.3\textwidth]{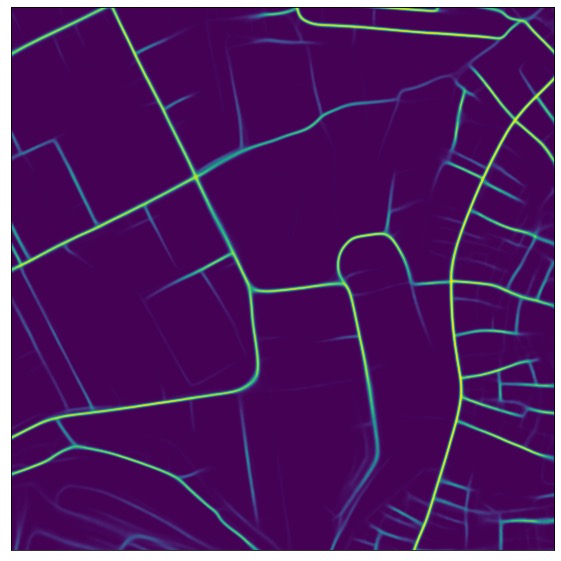} \\
  
  \end{tabular}
    \begin{tabular}{m{0.07\textwidth}C{0.3\textwidth}C{0.3\textwidth}C{0.3\textwidth}}
         & \footnotesize 50 cm RGB  & \footnotesize Building detection & \footnotesize Road detection  \\
    \end{tabular}

  \caption{Urban growth in Namanve industrial area, Kampala over time. Leftmost column is high-resolution imagery for reference.  }
\end{figure}

\begin{figure}
  \centering
  \begin{tabular}{m{0.07\textwidth}m{0.3\textwidth}m{0.3\textwidth}m{0.3\textwidth}}
  \scriptsize 2016-01-31 & \includegraphics[width=0.3\textwidth]{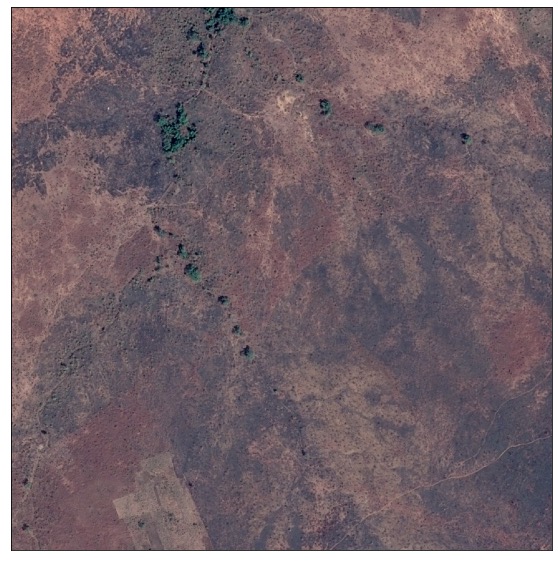} & \includegraphics[width=0.3\textwidth]{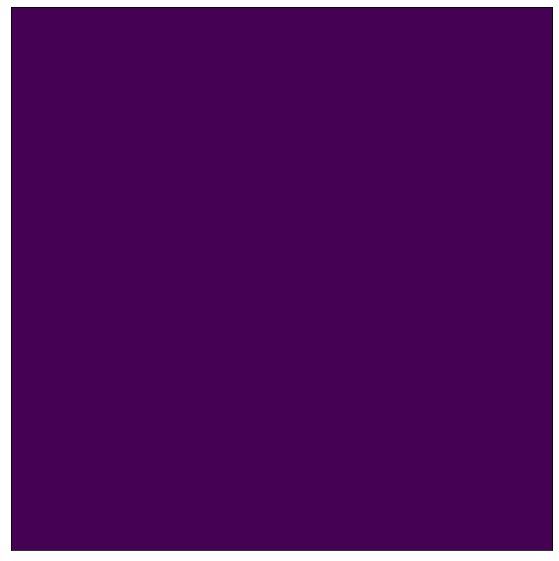} & \includegraphics[width=0.3\textwidth]{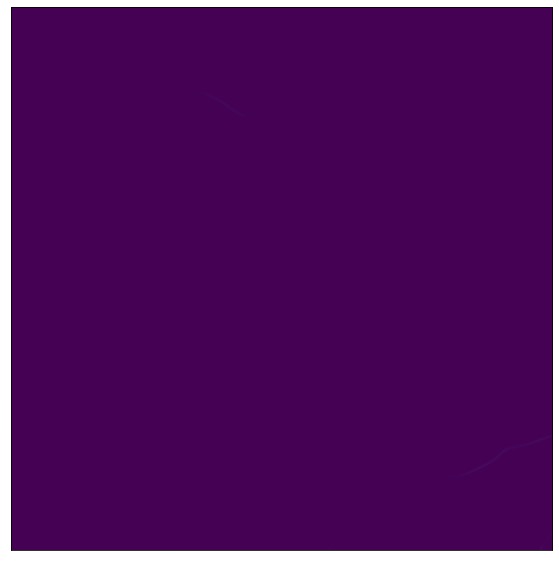} \\
  \scriptsize 2016-07-15 & \includegraphics[width=0.3\textwidth]{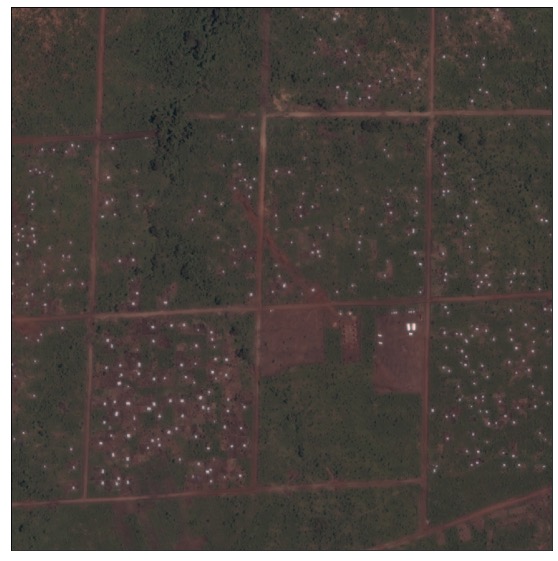} & \includegraphics[width=0.3\textwidth]{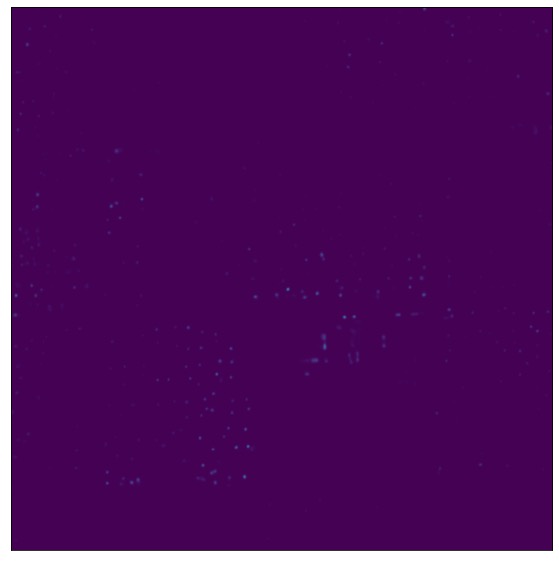} & \includegraphics[width=0.3\textwidth]{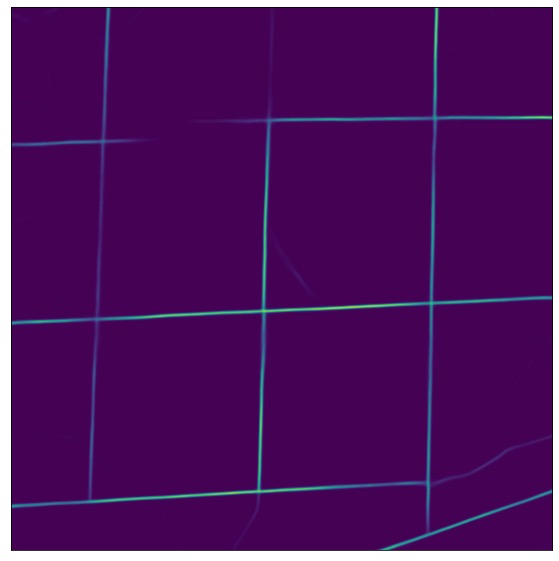} \\
  \scriptsize 2017-02-28 & \includegraphics[width=0.3\textwidth]{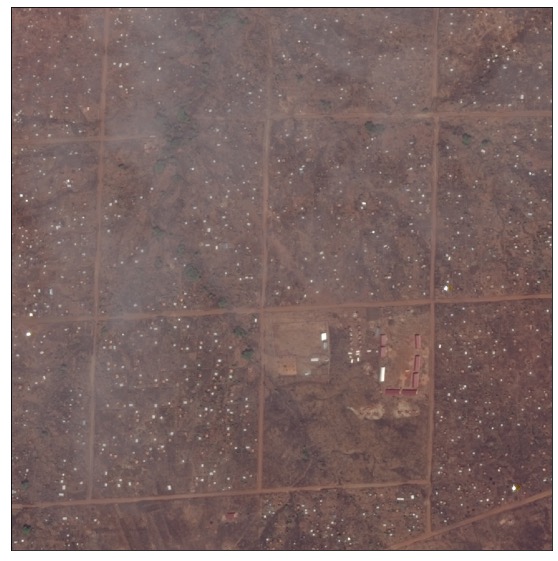} & \includegraphics[width=0.3\textwidth]{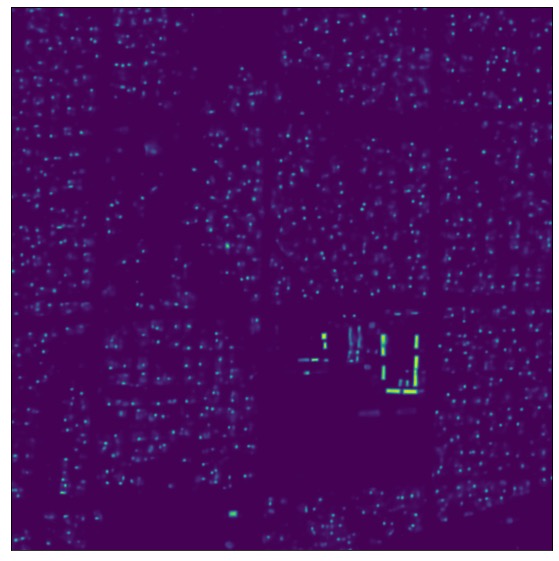} & \includegraphics[width=0.3\textwidth]{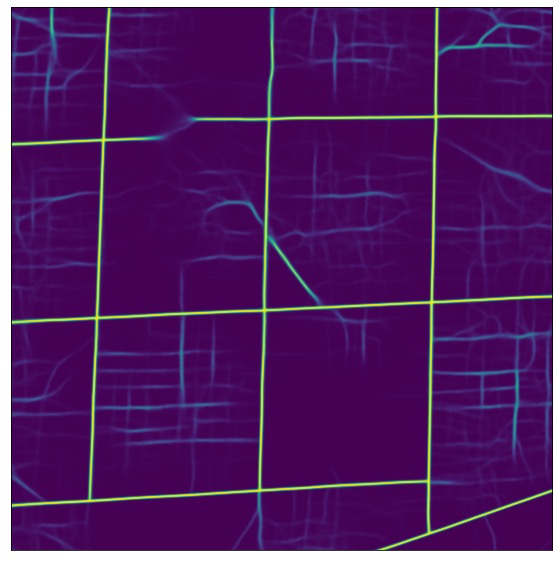} \\
  
  \end{tabular}
    \begin{tabular}{m{0.07\textwidth}C{0.3\textwidth}C{0.3\textwidth}C{0.3\textwidth}}
         & \footnotesize 50 cm RGB  & \footnotesize Building detection & \footnotesize Road detection  \\
    \end{tabular}

  \caption{The construction of Pagirinya refugee settlement in Eastern Adjumani District, Uganda. Leftmost column is high-resolution imagery for reference.}
\end{figure}

\end{document}